\PassOptionsToPackage{table}{xcolor}
\documentclass[sigconf]{acmart}

\usepackage{amsmath,amsfonts,bm}

\def\eqref#1{equation~\ref{#1}}

\def\1{\bm{1}}

\def\vf{{\bm{f}}}

\def\vp{{\bm{p}}}
\def\vq{{\bm{q}}}

\def\vv{{\bm{v}}}
\def\vw{{\bm{w}}}
\def\vx{{\bm{x}}}

\def\mA{{\bm{A}}}
\def\mB{{\bm{B}}}

\def\mE{{\bm{E}}}
\def\mF{{\bm{F}}}

\def\mK{{\bm{K}}}

\def\mQ{{\bm{Q}}}
\def\mR{{\bm{R}}}

\def\mV{{\bm{V}}}
\def\mW{{\bm{W}}}
\def\mX{{\bm{X}}}

\DeclareMathAlphabet{\mathsfit}{\encodingdefault}{\sfdefault}{m}{sl}
\SetMathAlphabet{\mathsfit}{bold}{\encodingdefault}{\sfdefault}{bx}{n}

\def\sQ{{\mathbb{Q}}}

\usepackage{multirow}
\usepackage{makecell} 
\usepackage{amsfonts}
\usepackage{subfig} 
\usepackage{subcaption}
\usepackage{enumitem}
\usepackage{makecell}
\usepackage{balance} 
\usepackage{hyperref}

\AtBeginDocument{%
  }

\copyrightyear{2025}
\acmYear{2025}
\setcopyright{acmlicensed}
\acmConference[SIGIR '25] {Proceedings of the 48th International ACM SIGIR Conference on Research and Development in Information Retrieval}{ July 13--18, 2025}{Padua, Italy.}
\acmBooktitle{Proceedings of the 48th International ACM SIGIR Conference on Research and Development in Information Retrieval (SIGIR '25), July 13--18, 2025, Padua, Italy}
\acmISBN{979-8-4007-1592-1/25/07}
\acmDOI{10.1145/XXXXXX.XXXXXX}

\begin{document}

\title{Continual Text-to-Video Retrieval with Frame Fusion and Task-Aware Routing}

\author{Zecheng Zhao}
\affiliation{%
  \institution{The University of Queensland}
  \city{Brisbane}
  \country{Australia}}
\email{uqzzha35@uq.edu.au}

\author{Zhi Chen}
\affiliation{%
  \institution{The University of Southern Queensland}
  \city{Toowoomba}
  \country{Australia}}
\email{zhi.chen@unisq.edu.au}

\author{Zi Huang}
\affiliation{%
  \institution{The University of Queensland}
  \city{Brisbane}
  \country{Australia}}
\email{huang@itee.uq.edu.au}

\author{Shazia Sadiq}
\affiliation{%
  \institution{The University of Queensland}
  \city{Brisbane}
  \country{Australia}}
\email{shazia@eecs.uq.edu.au}

\author{Tong Chen}
\authornote{Corresponding author.}
\affiliation{%
  \institution{The University of Queensland}
  \city{Brisbane}
  \country{Australia}}
\email{tong.chen@uq.edu.au}
\begin{abstract}
Text-to-Video Retrieval (TVR) aims to retrieve relevant videos based on textual queries. However, as video content evolves continuously, adapting TVR systems to new data remains a critical yet under-explored challenge. In this paper, we introduce the first benchmark for Continual Text-to-Video Retrieval (CTVR) to address the limitations of existing approaches. Current Pre-Trained Model (PTM)-based TVR methods struggle with maintaining model plasticity when adapting to new tasks, while existing Continual Learning (CL) methods suffer from catastrophic forgetting, leading to semantic misalignment between historical queries and stored video features. To address these two challenges, we propose FrameFusionMoE, a novel CTVR framework that comprises two key components: (1) the Frame Fusion Adapter (FFA), which captures temporal video dynamics while preserving model plasticity, and (2) the Task-Aware Mixture-of-Experts (TAME), which ensures consistent semantic alignment between queries across tasks and the stored video features. Thus, FrameFusionMoE enables effective adaptation to new video content while preserving historical text-video relevance to mitigate catastrophic forgetting. We comprehensively evaluate FrameFusionMoE on two benchmark datasets under various task settings. Results demonstrate that FrameFusionMoE outperforms existing CL and TVR methods, achieving superior retrieval performance with minimal degradation on earlier tasks when handling continuous video streams. Our code is available at: \url{https://github.com/JasonCodeMaker/CTVR}.

\end{abstract}



\maketitle
\begin{figure}[!t]
    \centering
    \includegraphics[width=1.0\linewidth]{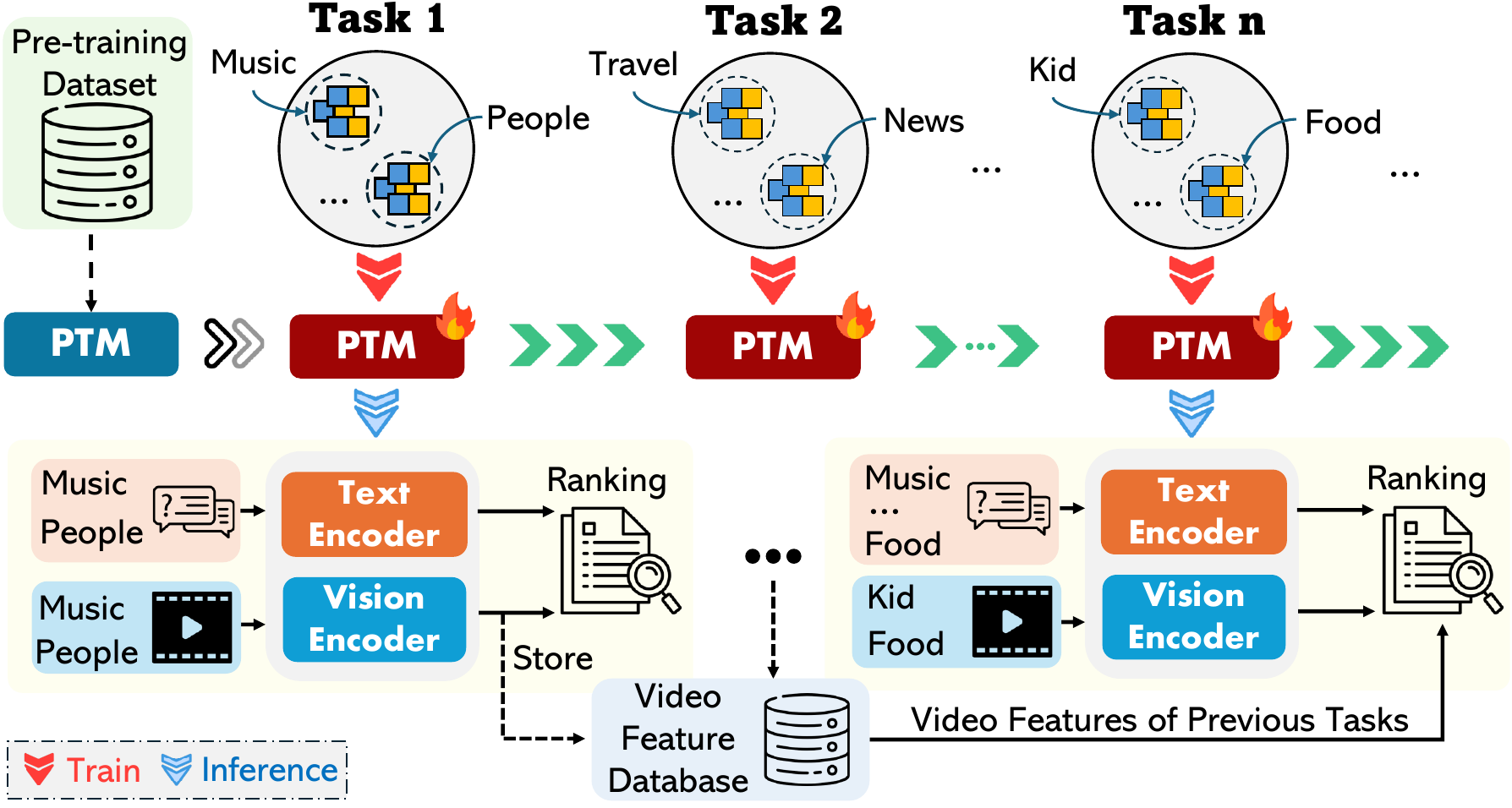}
    \Description{}
    \caption{An illustration of Continual Text-to-Video Retrieval (CTVR) pipeline. A Pre-Trained Model (PTM) continuously adapts to a sequence of TVR tasks through continual learning. Video features extracted in the current task are stored in a database and leveraged for subsequent tasks. During inference, all task queries can retrieve relevant videos within the video feature database. }
    \label{fig:intro}
    \vspace{-10pt}
\end{figure}

\section{Introduction}

The rapid growth of video-sharing platforms like YouTube has led to billions of Text-to-Video Retrieval (TVR) queries being processed daily \cite{gupta_youtube_searches}. With millions of new videos uploaded each day, these platforms continuously reflect evolving trends and shifting user interests. Meanwhile, the fast-paced content generation poses a unique challenge for TVR systems \cite{wu2023cap4video_tvr, fang2021clip2video_tvr, zhao2022centerclip_tvr, he2021improving_tra_tvr, jin2023diffusionret_tvr}: Continuous changes in data distribution make maintaining performance over time difficult for models. A naive solution of retraining TVR models with all the accumulated data is computationally expensive and difficult to scale. This raises a fundamental question: how can TVR systems adapt to new content over time without relying on historical data?

Continual Learning (CL) \cite{hadsell2020embracing, lopez2017gradient_cl, shin2017continual_cl, riemer2018learning_cl} offers a promising solution for sequential tasks by enabling models to learn new tasks without forgetting previously acquired knowledge. Motivated by real-world challenges in dynamic video retrieval, we explore the application of CL to tackle the critical and underexplored problem of Continual Text-to-Video Retrieval (CTVR) as shown in Figure \textcolor{red}{\ref{fig:intro}}. 
In practical scenarios of video-sharing platforms, the dynamic interests will continuously drive new video categories (\textit{e.g.,} trending topics) along with corresponding text queries, where the text-video pairs form distinct tasks with varying distributions over time. 
Following standard practices in industry applications \cite{covington2016deep,johnson2019billion}, we resort to an offline computation strategy for generating video features for each task, while processing the text queries in real time. This approach is necessitated as (1) performing real-time inference for each video is computationally prohibitive, and (2) text queries, being highly dynamic and user-driven, can be efficiently processed on-the-fly using a text encoder.
Under this setting, a CTVR model must learn to retrieve videos from these new categories using only current data, while ensuring it can retrieve videos across all previously learned categories during inference. 

\begin{figure}[t!]
    \centering
    \vspace{-10pt}
    \definecolor{red}{RGB}{255,150,150}
    \definecolor{lightgreen}{RGB}{150,200,150}
    \includegraphics[width=\linewidth]{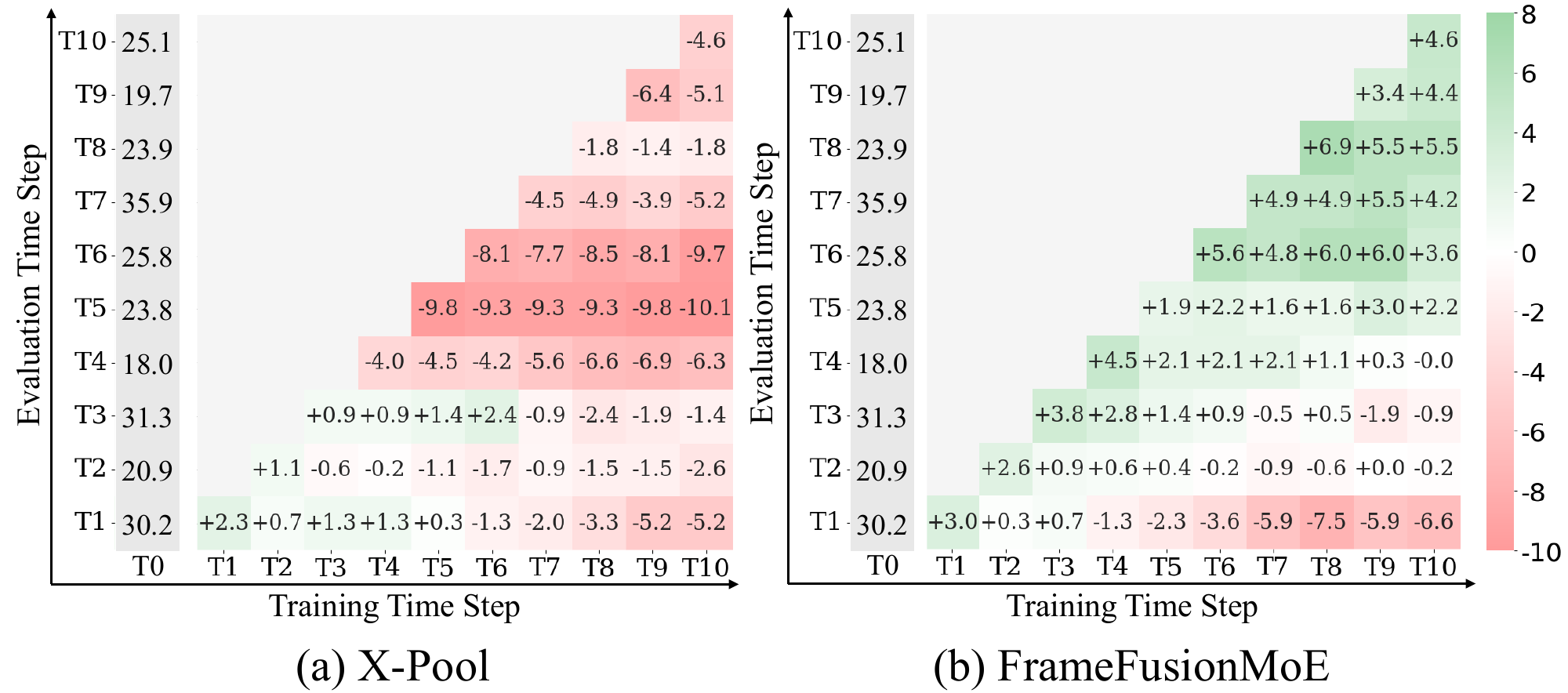}
    \Description{}
    \caption{Visualization of model plasticity across sequential tasks (T), indexed chronologically. The first column T0 denotes the initial state of the pre-trained model without any updates.  
    The presented results are performance variation on previous tasks after training on the current task (\textcolor{lightgreen}{green} for increase/\textcolor{red}{red} for drop) compared with the CLIP zero-shot results on MSRVTT dataset. (a) The state-of-the-art TVR method X-Pool \cite{{gorti2022XPool}} exhibits declining plasticity to new tasks, \text{i.e.,} underperform the zero-shot performance on the late stage tasks. (b) Our approach consistently improves task-wise performance while maintaining low backward forgetting when adapting to new tasks.}
    \vspace{-15pt}
    \label{fig:plasticity_loss}
\end{figure}

\begin{figure}[!htbp]
    \centering
    \vspace{-10pt}
    \definecolor{darkpurple}{RGB}{161, 134, 161}
    \definecolor{lightpurple}{RGB}{231, 192, 177}
    \definecolor{darkblue}{RGB}{83, 102, 140}
    \includegraphics[width=0.99\linewidth]{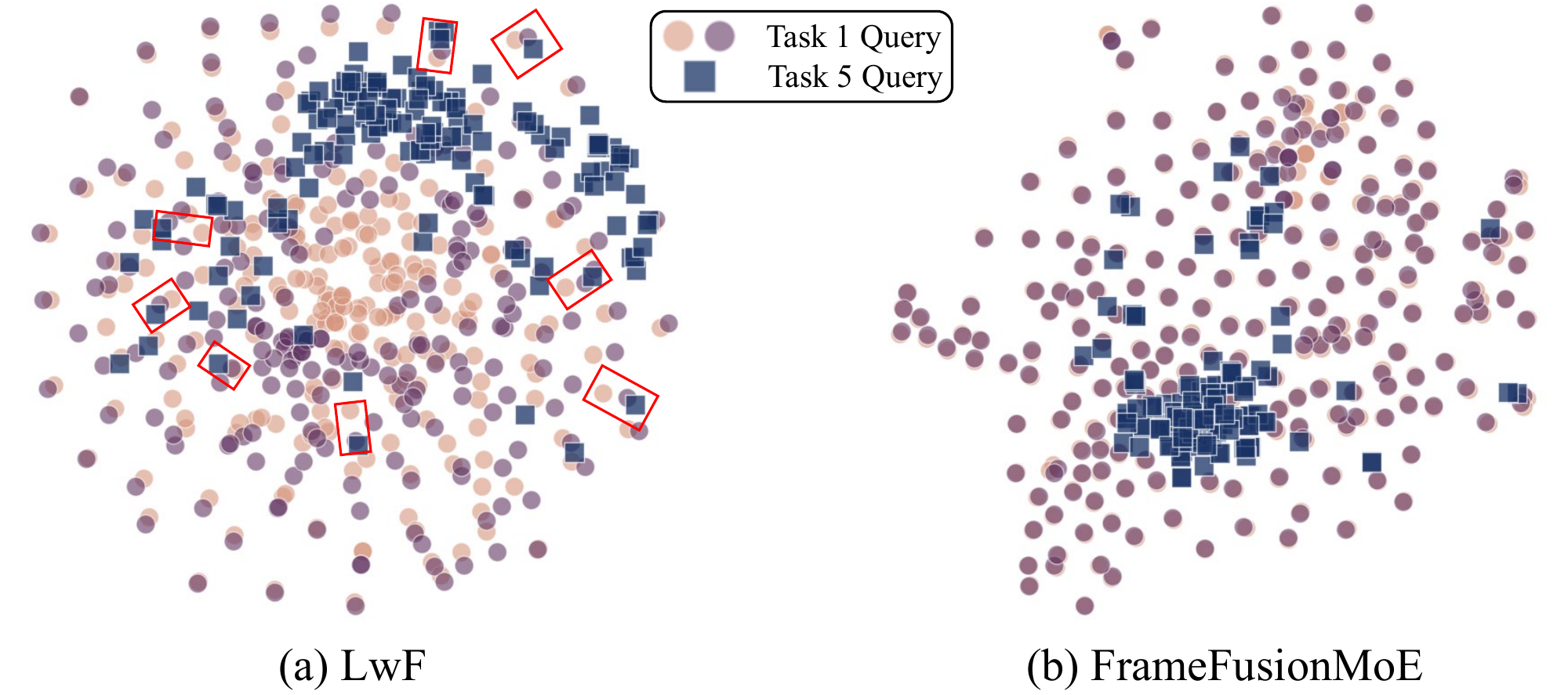}
    \Description{}
    \caption{
    The catastrophic forgetting problem, illustrated by text feature shifts on MSRVTT. Each \textcolor{darkpurple}{\huge $\bullet$} or \textcolor{darkblue}{\LARGE $\blacksquare$} represents a query in Task 1 or Task 5, respectively. In addition, we use different colors to mark the states of Task 1 queries after each task update. Ideally, if there is no forgetting at all, each Task 1 query should have no movements in the embedding space after learning new tasks. (a) LwF \cite{li2017LwF}, a strong CL baseline  shows the query embeddings shift from the original position while model keeps updating, as highlighted by the scattered colors \textcolor{red}{\LARGE $\Box$}. (b) Our approach maintains stable features, with minimal shifts across tasks, as evidenced by the overlap among different colors.}
    \vspace{-15pt}
    \label{fig:shift}
\end{figure}

Despite advancements in CL \cite{lomonaco2017core50,lin2021clear,srinivasan2022climb,villa2022vclimb,Tang2024vilcobench} and TVR methods \cite{luo2022clip4clip, gorti2022XPool,xue2023clipvip}, CTVR poses unique challenges to these approaches.
\textbf{(1) Model Plasticity Loss.} 
Existing TVR methods often fine-tune Pre-Trained Models (PTMs), such as CLIP \cite{radford2021CLIP}, to improve video-text alignment. However, such extensive modifications to the joint visual-semantic embedding space compromise the model’s plasticity, reducing its ability to generalize to future tasks. For instance, X-Pool \cite{gorti2022XPool} introduces additional networks for learning joint video-text embeddings, while CLIP-ViP \cite{xue2023clipvip} proposes video proxies to account for temporal relationships. Taking X-Pool as an example in Figure \textcolor{red}{\ref{fig:plasticity_loss}}(a), these methods require updating the entire model, which risks degrading transferability to future tasks as a result of overfitting to limited data early in training \cite{lyle2023understanding}.
\textbf{(2) Catastrophic Forgetting \cite{mcclellcatastrophic, french1999catastrophic, mccloskey1989catastrophic}:} 
Most existing CL methods are designed for recognition tasks, where the goal is to classify inputs into predefined categories. These methods focus on making task-specific representations discriminative but largely ignore the need to maintain stable embedding correlations across tasks. Unlike recognition tasks, retrieval demands consistent alignment between text queries and video features over time. In CTVR, historical videos are stored as embeddings, requiring the text encoder to stay aligned with these embeddings when adapting to new tasks, such that text queries from earlier tasks can still be correctly mapped to relevant videos. However, as shown in Figure \textcolor{red}{\ref{fig:shift}}(a), current CL methods fail to address this requirement, \textit{i.e.,} the same text queries {exhibit representation drift} in the feature space, {leading to overlapping query features across different tasks.} This shows signs of catastrophic forgetting on the alignment between historical queries and stored video features, which degrades retrieval performance.

To address the challenges in CTVR, we propose FrameFusionMoE, a parameter-efficient continual learning framework that tackles both model plasticity loss and catastrophic forgetting. FrameFusionMoE comprises two core components: (1) the \underline{F}rame \underline{F}usion \underline{A}dapter (\textbf{FFA}) to capture temporal video dynamics and maintain frame-wise representations, and (2) the \underline{T}ask-\underline{A}ware \underline{M}ixture-of-\underline{E}xperts (\textbf{TAME}) to route textual queries to task-specific experts for mitigating {query representation drift} across tasks.

In FrameFusionMoE, FFA is designed to preserve the semantics of image-text feature space of CLIP, thereby maintaining the model’s plasticity and transferability to future tasks. Our preliminary experiments indicate that composing video features by simply averaging frame features retains generalization across tasks. However, this approach neglects the temporal relationships inherent in video frame sequences. To address this limitation, we propose to propagate frame features sequentially while ensuring that frame-wise features remain intact. Specifically, we integrate the FFA into each Transformer block of the CLIP image encoder, where each FFA is implemented with a multi-head cross attention unit that takes previous frame features as the query and current frame features as the key and the value. As reflected by Figure \textcolor{red}{\ref{fig:plasticity_loss}}(b), this enables the model to effectively capture temporal dependencies without compromising the generalizability of the joint image-text feature space of CLIP.

To tackle the catastrophic forgetting on the misalignment between historical queries and the stored video features, we introduce TAME. It preserves previously learned representations while allowing the model to adapt to new tasks by selectively routing text queries to the appropriate experts. The architecture is built on a Mixture-of-Experts (MoE) design, where each expert is implemented using LoRA-like layers \cite{hulora, tian2024hydralora}. To achieve task-specific routing, we propose a task-aware router that constructs task prototypes, which guide the model in selecting the most relevant expert for handling task-dependent queries. This design minimizes representation drift by ensuring that embeddings from previous tasks remain aligned with their corresponding queries across time.

Overall, the contributions of this paper are:
\begin{itemize}[leftmargin=*] 
\item We propose a FrameFusionMoE framework for Continual Text-to-Video Retrieval (CTVR), a practical yet under-explored area of video understanding. \textbf{To the best of our knowledge, FrameFusionMoE is the first attempt for CTVR.}
\item We design two novel  components in FrameFusionMoE to address the unique challenges in CTVR: Frame-Fusion Adapter (FFA) and Task-Aware Mixture of Experts (TAME). The FFA preserves the image-text embedding space of Pre-Trained Models (PTMs) while capturing temporal video dynamics, ensuring model plasticity for future tasks. TAME mitigates catastrophic forgetting by using a task-aware routing mechanism to maintain consistent alignment between text queries and stored video features, ensuring long-term retrieval performance.
\item  We benchmark CTVR on two text-to-video datasets, \textit{i.e.,} MSRVTT \cite{xu2016msrvtt} and ActivityNet \cite{caba2015activitynet}. We re-purpose and evaluate four state-of-the-art continual learning (CL) and three text-to-video retrieval (TVR) baselines for CTVR. Our extensive benchmarking results highlight the limitations of existing methods and verify the advantageous performance of FrameFusionMoE. 
\end{itemize}

\vspace{-0.2cm}
\section{Related Work}
\noindent\textbf{Text-to-Video Retrieval}
Deep learning has revolutionized computer vision \cite{chen2020canzsl_cv,ho2020denoising_cv,chen2024fastedit_cv,chen2021semantics_cv,chen2020rethinking_cv,vaswani2017attention_cv}, creating a foundation for advances in Text-to-Video Retrieval (TVR) \cite{wu2023cap4video_tvr, fang2021clip2video_tvr, zhao2022centerclip_tvr, jin2023diffusionret_tvr, fang2023uatvr, yang2024dgl_tvr, guan2023pidro_tvr, li2023progressive_tvr} through pre-trained Vision-Language Models like CLIP \cite{radford2021CLIP}, which bridge textual descriptions and visual content. CLIP4Clip \cite{luo2022clip4clip} pioneered the application of CLIP models for text-to-video retrieval tasks, demonstrating CLIP's robust transfer learning capabilities. However, recognizing that videos contain unique temporal information that static images lack, there is a significant domain gap between videos and images.  Many methods \cite{luo2022clip4clip, gorti2022XPool, xue2023clipvip, liu2022TS2-Net, deng2023promptswitch, wang2024textmass, wu2023cap4video, fang2021clip2video} have been proposed to leverage temporal dynamics to enhance video representations. X-Pool  \cite{gorti2022XPool} leverages text-conditioned feature attention across video frames to generate semantically enriched embeddings. Meanwhile, TS2-Net \cite{liu2022TS2-Net}, CLIP-ViP \cite{xue2023clipvip}, and Prompt Switch \cite{deng2023promptswitch} enhance video representations by incorporating temporal or video-specific embeddings to capture inter-frame relationships. Unlike video representation learning, T-MASS \cite{wang2024textmass} utilizes stochastic text embeddings to strengthen video-text alignment. Although these architectural modifications to CLIP enable effective adaptation to video-language alignment, they compromise CLIP's original generalization capabilities. Our proposed Frame Fusion Adapter (FFA) enhances temporal dynamics of video frames while preserving model plasticity for new tasks.

\begin{figure*}[!htp]
    \vspace{-0.2cm}
    \centering
    \includegraphics[width=1.0\linewidth]{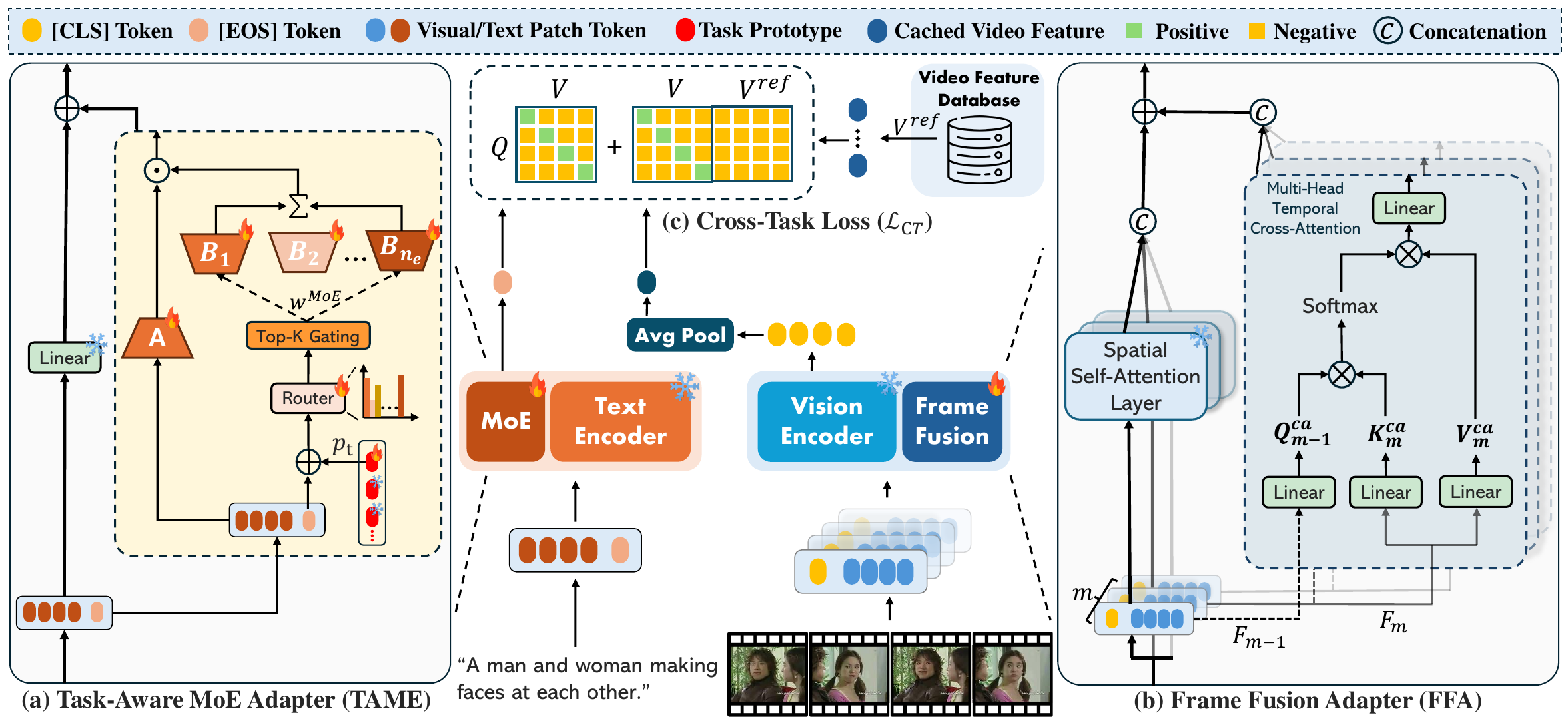}
    \vspace{-0.4cm}
    \Description{}
    \caption{Overall framework of the FrameFusionMoE.  It consists of three core components: (a) A Task-Aware MoE Adapter (TAME) that is added to a frozen CLIP text encoder to learn the distribution of text query through the selection of multiple experts $\{\mB_i\}_{i=1}^{n_e}$. The expert weights $\vw^\text{MoE}$ are determined by a router taking the element-wise addition ($\oplus$) of the $[\text{EOS}]$ token and task prototype $\vp_t$ as input. (b) A vision processing pipeline where frame features are processed through a frozen CLIP vision encoder and Frame Fusion Adapters (FFA). Each FFA uses previous frame feature maps $\mF_{m-1}$ to attend over current frame $\mF_m$ through multi-head temporal cross attention. The FFA output serves as a temporal guidance signal that is added back to each spatial self-attention layer. (C) The Cross-Task Loss ($\mathcal{L}_{\text{CT}}$) optimizes representations by drawing matched text-video pairs closer while pushing away cached video features that serve as negative samples. }
    \label{fig:Model_Architecture}
    \vspace{-0.3cm}
\end{figure*}

\vspace{5pt}
\noindent\textbf{Continual Learning.}
Continual Learning (CL) \cite{hadsell2020embracing, lopez2017gradient_cl, shin2017continual_cl, riemer2018learning_cl, QiuYHC20_CL, GuoYWCZH19_CL, YuanYHCWC22_CL} is a machine learning paradigm where models learn sequentially from a stream of tasks while maintaining performance on previous tasks. Class Incremental Learning (CIL) \cite{hsu2018CIL, van2019CIL, rebuffi2017icarl_cil} represents a challenging paradigm where each task contains non-overlapping categories, and test samples may come from any previous task, requiring models to balance stability and plasticity. Recently, some task-specific benchmarks \cite{roady2020stream, garg2023tic, lomonaco2017core50, lin2021clear, srinivasan2022climb, villa2022vclimb, Tang2024vilcobench} have been developed that enable standardized evaluation protocols between different CL methods. In image domain, Core50 \cite{lomonaco2017core50}, CLeAR \cite{lin2021clear} and CLiMB \cite{srinivasan2022climb} compare and analyze various CL methods from different perspectives. Extending to the video domain, vCLIMB \cite{villa2022vclimb} introduces the first benchmark for continual learning in video action recognition, while ViLCo-Bench \cite{Tang2024vilcobench} subsequently focused on evaluating continual learning tasks in the video-language domain. However, Text-to-Video Retrieval (TVR) remains underexplored despite its practical importance. In this work, we benchmark continual learning for text-to-video retrieval.

\vspace{5pt}
\noindent\textbf{Continual Learning with Pre-Trained Models.}
In traditional CL approaches, models learn incrementally from sequential tasks, often leading to overfitting on the initial task. Fortunately, with the advent of large-scale pre-trained models (PTMs), the field has shifted its focus towards leveraging their robust representation capabilities \cite{zhou2024revisiting_ptm_cl, mcdonnell2024ranpac_ptm_cl, zhang2023slca_ptm_cl}. These emerging approaches can be categorized into three main paradigms: prompt-based methods \cite{wang2022L2P, wang2022dualprompt, jung2023DAP, prompt_cl, wang2022sprompt, nicolas2023mop_prompt}, regularization-based methods \cite{li2017LwF, ding2022VR_LwF, zheng2023ZSCL}, and model mixture-based methods \cite{wortsman2022wise, yu2024MoEAdapter, zhou2023learning_mm, wang2023isolation_mm, chen2025promptfusion_mm}. The prompt-based methods leverage the strong generalization capabilities of PTMs by introducing minimal trainable prompt parameters, enabling efficient adaptation. L2P \cite{wang2022L2P} maintains a prompt pool and selects the most relevant prompts for each test sample. DualPrompt \cite{wang2022dualprompt} incorporates additional task-specific prompts, enabling the model to encode both task-invariant patterns and task-specific instructions. Moreover, PIVOT \cite{PIVOT_villa} integrates prompting mechanisms into video CL for adaptive prompt selection. Regularization-based methods introduce regularization terms to achieve a balance between stability and plasticity. EWC \cite{kirkpatrick2017EWC}, SI \cite{zenke2017SI} and MAS \cite{aljundi2018MAS} employ parameter-specific regularization terms that add a penalty to the weight updates based on each parameter's importance for previously learned tasks. On the other hand, LwF \cite{li2017LwF} mitigates catastrophic forgetting with knowledge distillation by treating the previous model as a teacher and the current model as a student.  LwF-VR \cite{ding2022VR_LwF} utilizes the CLIP vocabulary set as a reference, while ZSCL \cite{zheng2023ZSCL} leverages ImageNet as a reference dataset, both aiming to better preserve the pre-trained model's capabilities. In contrast to training-phase optimization, model mixture-based methods tackle catastrophic forgetting during inference by combining experts from different tasks. 
MoE-Adapter \cite{yu2024MoEAdapter} incorporates Mixture-of-Experts (MoE) \cite{fedus2022switchMoE, jacobs1991adaptiveMoE, shazeer2017outrageouslMoE} as specialized adapters. Each expert is trained to handle distinct knowledge distributions. However, TVR poses a unique catastrophic forgetting challenge on the alignment between historical queries and stored video features. To cope with such unique challenge, our proposed task-aware mixture-of-experts can maintain the historic queries distribution while adapting to new tasks.

\section{Continual Text-to-Video Retrieval}
In this section, we first formalize the practical yet under-studied research problem of learning a text-to-video retrieval system in sequential tasks, \textit{i.e.,} Continual Text-to-Video Retrieval (CTVR). Then, we discuss the motivation of the proposed method. Lastly, we introduce the components of our proposed FrameFusionMoE.

\subsection{Problem Definition} 

In CTVR, we aim to train a retrieval model that allows text queries $q$ to retrieve relevant videos $v$ while videos of new tasks emerge. 
Given a sequence of $T$ tasks $\{D_1, \cdots, D_T\}$, where each task $D_t = \{(q^{t}_i, v_i^t)\}_{i=1}^{n_t}$ contains $n_t$ query-video pairs from categories $C_t$, where $C_i \cap C_j = \emptyset, \forall i \neq j$. 
\begin{sloppy}
During training on task $t$, the data access is restricted to only $D_t$, while the data of all previous tasks $\{D_1,\cdots,D_{t-1}\}$ are inaccessible. 
For each query-video pair, we uniformly sample $M$ frames from a video $v = [f_1, f_2,\cdots,f_M]$, each is extracted as frame features $\vv = [\vf_1, \vf_2, \cdots, \vf_M]$ by the CLIP image encoder. $\vf_m$ represents the [\texttt{CLS}] token from the Transformer-based vision-encoder for $m$-th frame. The text encoder takes the [\texttt{EOS}] token as query features $\vq$.
After training on a task $t$, the testing video features $\mV_t = \{ \vv_i \}^{n_t}_{i=1}$ of categories $C_t$ are saved into the database $\mV_{[1:t]} = \mV_{[1:t-1]} \bigcup \mV_t$.
\end{sloppy}
During testing at task $t$, given a set of test queries $\sQ = \bigcup_{i=1}^t \sQ_i$ from all previous tasks, the model retrieves relevant videos from the database $\mV$. For each query $q \in \sQ$, the retrieval is performed by computing the cosine similarity $sim(\vq, \vv))$ between the query features $\vq$ and the video features $\vv$.
The videos are then ranked according to $\text{rank}(\vv) = \text{sort}_{\vv \in \mV_{[1:t]}}(\text{sim}(\vq,\vv))$ in descending order.

\vspace{-6pt}
\subsection{Motivation}\label{sec:challenges} 
To explore the key aspects of an effective CTVR system, we analyze state-of-the-art methods for TVR and CL, so as to locate the research challenges when two areas intersect. 

Recent TVR methods primarily adapt the knowledge of Pre-Trained Models (PTMs) from the image-text domain to the video-text domain. 
These adaptations often require significant modifications to the joint image-text embedding space to account for video temporality.
For instance, X-Pool \cite{gorti2022XPool} introduces additional networks for learning joint video-text embeddings. CLIP-ViP \cite{xue2023clipvip} incorporates video proxy tokens to account for temporal relationships.
However, as shown in Table \ref{tab:Overview}, the TVR baselines underperform the Average Pooling baseline \cite{luo2022clip4clip} which simply averages frame features to represent video features. We hypothesize that the adaptations are designed as one-off solutions, focusing solely on the current task while failing to retain the pretraining knowledge of PTMs. This results in \textbf{model plasticity loss}, hindering the model's ability to adapt to future tasks. 

In CTVR, re-extracting video features from historical data after each task is computationally prohibitive. A feasible alternative is to cache historical video features into a database after each task. Thus, an effective CTVR system must maintain semantic alignment between historical queries and these cached video features while learning new tasks. Unfortunately, existing CL methods are predominantly designed for recognition tasks, where the focus is on classifying inputs into predefined categories. These methods prioritize task-specific discriminative features but neglect the need to maintain stable text query features across tasks. 
For example, Mixture-of-Experts Adapter (MoE-Adapter) \cite{yu2024MoEAdapter} has demonstrated continual adaptation to sequential tasks, where the experts are dynamically updated for new tasks.
This leads to \textbf{catastrophic forgetting}, where the alignment between historical queries and stored video features deteriorates.

Our framework consists of: (1) the Frame Fusion Adapter (FFA) that captures video temporality while preserving CLIP’s image-text embedding space, ensuring model plasticity for future tasks; and (2) the Task-Aware Mixture-of-Experts (TAME) that maintains alignment between historical queries and cached video features, mitigating catastrophic forgetting in the embedding space. In what follows, we unfold the design of those core components. 

\vspace{-5pt}
\subsection{Frame Fusion Adapter}

The Frame Fusion Adapter (FFA) is designed to preserve CLIP's model plasticity while enabling the learning of video temporality in a parameter-efficient manner. Following the design principles of AvgPool \cite{luo2022clip4clip}, Frame Fusion maintains the joint image-text embedding space of CLIP without introducing disruptive modifications, thereby retaining the model's pre-trained generalization capabilities. To achieve this, FFAs are introduced as lightweight adapters that are placed between the Transformer blocks of CLIP, while keeping the PTM parameters frozen. This approach minimizes computational overhead while allowing the model to adapt to video-specific tasks. Furthermore, FFAs enable dependent frame features to propagate across frames, effectively capturing the temporal dynamics inherent in videos. 

As shown in Figure \textcolor{red}{\ref{fig:Model_Architecture}}(b), FFAs are implemented with Cross-Attention (CA) blocks in parallel with existing Self-Attention (SA) blocks in CLIP. Essentially, the input image tokens of SA are simultaneously fed into both SA and CA blocks, and the output from both are added together for the following Transformer blocks. Specifically, consider the image tokens of each frame $[\mF_{1}, \mF_{2}, \cdots, \mF_{M}]$, where $M$ is the number of frames. 
Then, for the $m$-th frame in a video, CA can be formulated as:
\begin{equation}
  \begin{aligned}
\mQ^{ca}_{{m-1}} = \mF_{\textcolor{red}{m-1}} \mW_q^T, \quad 
\mK^{ca}_{{m}} = \mF_{\textcolor{red}{m}} \mW_k^T,  \quad 
\mV^{ca}_{{m}} = \mF_{\textcolor{red}{m}} \mW_v^T, \quad \quad  
\label{eq:CA}
  \end{aligned}
\end{equation}
\begin{equation*}
  \begin{aligned}
\mA^{ca} = \texttt{softmax}(\mQ^{ca}_{m-1} \mK^{ca T}_{m} / \sqrt{O/h} ),~  CA(\mF_{\textcolor{red}{m-1}}, \mF_{\textcolor{red}{m}}) = \mA^{ca} \mV^{ca}, 
\label{eq:CA2}
  \end{aligned}
\end{equation*}
where the attention query comes from the patch tokens of the previous frame $\mF^{m-1}$ and the key/values are the current frame $\mF_{m}$. $\mW_q, \mW_k$ and $\mW_v \in \mathbb{R}^{O\times (O/h)}$ are trainable linear layers, $O$ and $h$ are the feature dimension and number of heads. The CA block is implemented in parallel to the existing SA blocks in CLIP. Similarly, SA blocks can be formulated as:
\vspace{-0.1cm}
\begin{equation}
  \begin{aligned}
\mQ^{sa}_{{m}} = \mF_{\textcolor{red}{m}} \tilde{\mW}_q^T, \quad\quad 
\mK^{sa}_{{m}} = \mF_{\textcolor{red}{m}} \tilde{\mW}_k^T,  \quad\quad 
\mV^{sa}_{{m}} = \mF_{\textcolor{red}{m}} \tilde{\mW}_v^T, \quad\\ 
\mA^{sa} = \texttt{softmax}(\mQ^{sa}_m \mK^{sa T}_{m} / \sqrt{O/h} ),\quad   SA(\mF_{\textcolor{red}{m}}, \mF_{\textcolor{red}{m}}) = \mA^{sa} \mV^{sa}, 
\label{eq:CA3}
  \end{aligned}
\end{equation}
\vspace{-0.1cm}
The fused tokens are added with the original tokens via residual connections:
\vspace{-0.1cm}
\begin{equation}
  \begin{aligned}
\mF_{\textcolor{red}{m*}} = SA( \mF_{\textcolor{red}{m}}, \mF_{\textcolor{red}{m}}) + \alpha~CA(\mF_{\textcolor{red}{m-1}}, \mF_{\textcolor{red}{m}})
  \end{aligned}
\end{equation}
\vspace{-0.1cm}
where $\alpha$ is a trainable parameter that controls the weight of CA.
The obtained the new frame tokens $\mF_{m*}$ is further fed into the following Transformer blocks and after each, we have an FFA module applied.

\begin{figure}[tbp]
    \centering
    \includegraphics[width=1.0\linewidth]{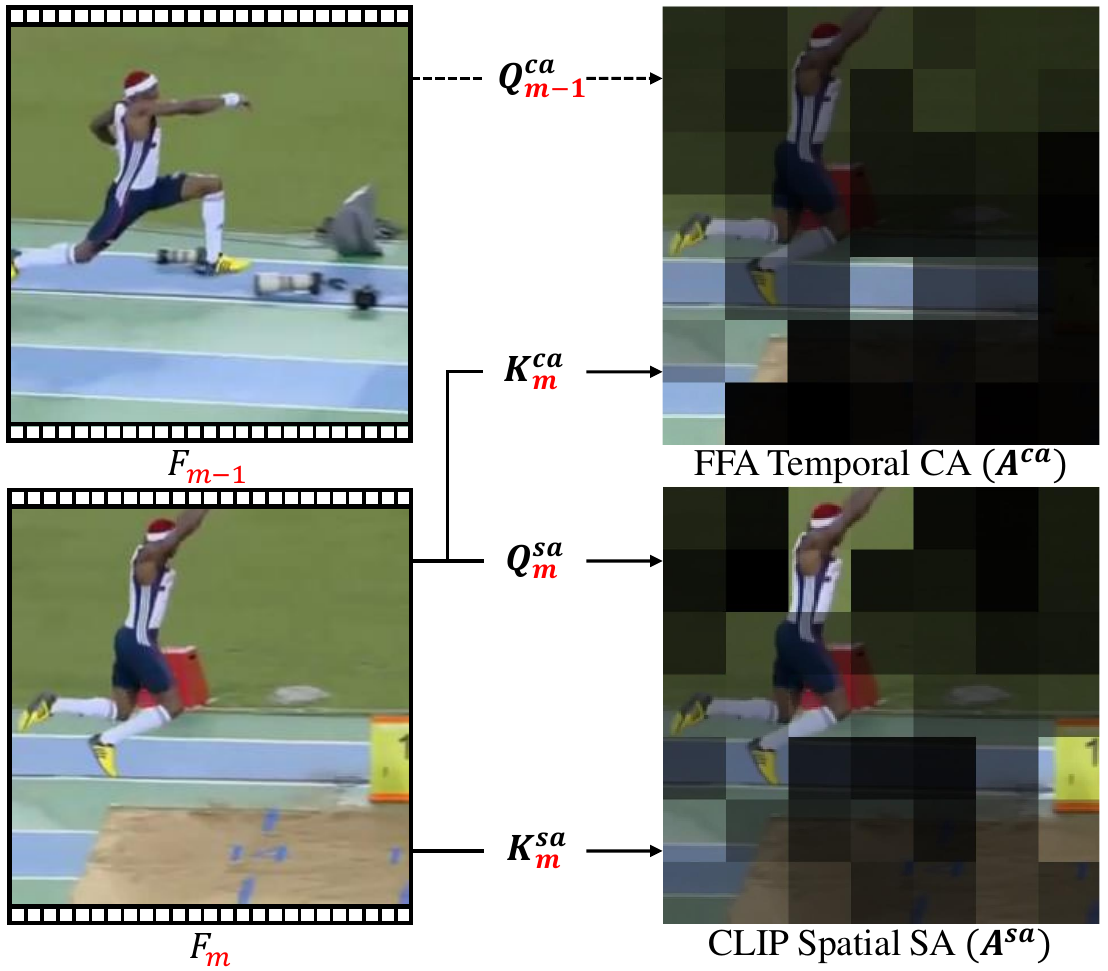}
    \vspace{-0.6cm}
    \Description{}
    \caption{Visualization of attention maps from FFA Temporal CA and CLIP Spatial SA mechanisms. Brighter regions in the attention maps indicate higher attention weights. FFA's temporal CA demonstrates stronger attention weights on temporally consistent regions between frames (e.g., track surface, background) while showing lower attention on the changing sand pit area, effectively capturing inter-frame consistency. CLIP's spatial SA focuses on the athlete and their jumping action, capturing semantically important motion information within the frame.}
    \label{fig:FFA_CA}
    \vspace{-0.4cm}
\end{figure}

The SA blocks primarily capture intra-frame patch relationships, while the CA blocks focus on inter-frame temporal dynamics by integrating features from adjacent frames. By combining these mechanisms, FFAs efficiently utilize frame-level features extracted at each CLIP layer to enrich video representations with temporal information. 
To gain further insights into the behavior of FFAs, we visualize the attention matrix $\mA^{ca}$ and $\mA^{sa}$. As shown in Figure \textcolor{red}{\ref{fig:FFA_CA}}, the CA attention query from the previous frame $\mQ^{ca}_{m-1}$, predominantly attends to temporally consistent regions across frames such as the track surface and background elements. This behavior ensures inter-frame consistency while de-emphasizing regions undergoing rapid changes, such as athlete's movement. In contract, the SA mechanism primarily focuses on the dynamic foreground elements, effectively capturing semantically important motion information within the frame. 
This combination of intra-frame and inter-frame attention mechanisms facilitates effective video representation learning through Frame Fusion while preserving the original architecture of CLIP. Consequently, the proposed approach ensures both computational efficiency and strong generalization across continual learning tasks.

\vspace{-0.2cm}
\subsection{Task-Aware Mixture-of-Experts Adapter}
To mitigate the catastrophic forgetting problem caused by misalignment between historical queries and stored video features, we introduce Task-Aware Mixture-of-Experts (TAME) adapters. The primary objective of TAME is to learn task-conditional text features, ensuring alignment across sequential tasks while maintaining efficient adaptation.

As illustrated in Figure \textcolor{red}{\ref{fig:Model_Architecture}}(a), TAME adapters are integrated with the linear layers in the SA blocks of the CLIP text encoder. A TAME adapter consists of a set of expert networks $\{\mE_i(\cdot)\}_{i=1}^{n_e}$ where $n_e$ represents the number of experts, a router function $\mR(\cdot)$, and task prototypes $\{\vp_t\}_{t=1}^{T}$.

Given an input textual query with $n_x$ tokens $\mX = [\vx_{1}, \vx_{2}, \cdots, \\ \vx_{n_x}, \vx^{[EOS]}]$, where $\vx^{[EOS]}$ is the end-of-sentence token representing the text features of the entire sentence. During task $t$, the token $\vx^{EOS}$ is added with the current task prototype $\vp_t$. The resulting features are passed into the router to produce the gating parameter $\vw^{MoE}$, which determines the activation of experts:
\vspace{-0.1cm}
\begin{equation}
  \begin{aligned}
  \vw^{{MoE}} = \texttt{softmax}(\text{TopK}(\mR(\vx^{[EOS]} + \vp_{t}))),
\label{eq:CA4}
  \end{aligned}
\end{equation}
\vspace{-0.1cm}
where $TopK$ are the experts with highest confidence. The selected $K$ experts will take the tokens $\mX$ as input. Specifically, the experts are parameterized using a LoRA structure, where the low-rank encoder $\mA$ is shared among experts while the expert-specific decoders are represented as $\{\mB_i\}_{i=1}^{n_e}$. The experts output is then formulated as:
\vspace{-0.1cm}
\begin{equation}
\mE(\mX) =\sum_{i \in \text{TopK}} \vw^{{MoE}} \cdot (\mathbf{B}_i\mathbf{A}\mX).
\label{eq:MoE}
\end{equation}
Let $\tilde{\mW}$ be the frozen layers in the SA blocks of CLIP text encoder, the output of the layer coupled with TAME is formulated as:
\begin{equation}
\mX^* = \mX \tilde{\mW}^{T} + \lambda~\mE(\mX),
\label{eq:MoE2}
\end{equation}
\vspace{-0.1cm}
where $\lambda$ controls the scale of the residual features. 

During inference, for a given textual query, each task prototype $\vp_{i} (\text{where} ~i \in \{1,\cdots,T\}$ is added to the end-of-token $\vx^{[EOS]}$, resulting in task-conditional text features $\{\vq_i\}^{T}_{i=1}$. These features are then used to compute similarity scores with the stored video features from different tasks $\mV_{[1,T]}$, and the $TopK$ most relevant videos are retrieved.

\vspace{-0.1cm}
\subsection{Optimization}
To effectively align representations between video and textual modalities based on a pre-trained CLIP model, we optimize cross-modal alignment in a shared embedding space for each task. They are video-to-text ($v2t$) and text-to-video ($t2v$) modalities:
\vspace{-0.1cm}
\begin{equation}
  \begin{aligned}
\mathcal{L}_{v2t} = -\mathbb{E}_{i}\log\frac{\exp(\langle \vq_i, \vv_i \rangle/\tau)}{\sum_{j}\exp(\langle \vq_j, \vv_i \rangle/\tau)}, \\ \mathcal{L}_{t2v} = -\mathbb{E}_{i}\log\frac{\exp(\langle \vq_i, \vv_i \rangle/\tau)}{\sum_{j}\exp(\langle \vq_i, \vv_j \rangle/\tau)}.
\label{eq:Info_NCE}
  \end{aligned}
\end{equation}
We also introduce a Cross-Task (CT) loss that leverages the samples within the video feature database as negative references. By considering videos from previous tasks, this loss performs semantic regularization on the alignment between queries and their relevant videos. It maintains task-specific feature distributions while preventing catastrophic forgetting. We formulate CT loss as:
\vspace{-0.1cm}
\begin{equation}
\mathcal{L}_{CT} = -\mathbb{E}_{i}\left[\log\frac{\exp(\langle \vq_i, \vv_i \rangle/\tau)}{\sum_{t}\exp(\langle \vq_i, \vv_t \rangle/\tau) + \sum_{h}\exp(\langle \vq_i, \vv^{ref}_h \rangle/\tau)}\right]
\label{eq:triplet}
\end{equation}
\vspace{-0.1cm} 
where $\vq_i$ and $\vv_i$ denote the text query and its matched video features from the current task, $\vv_t$ represents current task video features as in-batch negatives, $\vv^{ref}_h$ denotes video features of previous tasks serving as additional negative references, and $\tau$ is a temperature parameter.
The overall objective for cross-modal alignment between text and video representations is defined as:
\vspace{-0.1cm}
\begin{equation}
\mathcal{L} = (1 -\beta) \frac{1}{2}(\mathcal{L}_{v2t} + \mathcal{L}_{t2v}) + \beta\mathcal{L}_{CT},
\label{eq:final_loss}
\end{equation}
\vspace{-0.1cm}
where $\beta$ is a pre-defined hyper-parameter with $\beta = 0$ for task 1.

\vspace{-0.2cm}
\section{Benchmark Experimental Setup}
We first present the experimental setup for CTVR benchmark, including datasets (Section \ref{sec:datasets}), evaluation metrics (Section \ref{sec:eval}), baselines (Section \ref{sec:baselines}), and implementation details (Section \ref{sec:implementation}). Then, we conduct comprehensive experiments to address each research question in Section \ref{sec:exp}.


\vspace{-0.2cm}
\subsection{Datasets}
\label{sec:datasets}
We construct CTVR using two established TVR datasets with pre-defined categorical structures.
To comprehensively assess CL capabilities, we evaluate models across two settings with 10 and 20 sequential tasks.
\noindent \textbf{(1) MSRVTT} \cite{xu2016msrvtt} consists of 10,000 videos (10-32 seconds) and 200,000 captions across 20 distinct classes. Traditional TVR works utilize \textit{Train-7K} \cite{miech2019howto100m} for training and 1K-A test set \cite{yu2018joint} for evaluation. 
\textbf{(2) ActivityNet Captions} \cite{caba2015activitynet} consists of 20,000 video clips (average 180 seconds) and 100,000 descriptions across 200 categories. 
Different from traditional TVR evaluation pipelines that concatenate all descriptions for paragraph-video retrieval, we ultilize the trimmed subset \cite{caba2015activitynet} and employ LLMs \cite{dubey2024llama} to select the most category-representative description-video clip pair from each video. This single-pair selection better aligns with real-world search scenarios while maintaining retrieval complexity.

\begin{table*}[!t]
    \centering
    \definecolor{red}{RGB}{255,150,150}
    \definecolor{lightgreen}{RGB}{150,200,150}
    \definecolor{gray}{RGB}{128,128,128}
    \definecolor{lightpurple}{RGB}{248, 231, 242}
    \caption{Comparison of model performance for CTVR on MSRVTT and ACTNET datasets with 10 and 20 tasks, respectively. The top two models are highlighted in bold and underlined. `E.R.' means the utilization of an external reference dataset.}
    \vspace{-0.15cm}
    \setlength{\tabcolsep}{0.35pt}
    \footnotesize
    \scalebox{0.86}{\begin{tabular}{lccccccc|cccccc|cccccc|cccccc|c}
    \hline
    \multirow{3}{*}{Model} & \multirow{3}{*}{\makecell{E.R.}} & \multicolumn{6}{c}{MSRVTT-10} & \multicolumn{6}{c}{MSRVTT-20} & \multicolumn{6}{c}{ACTNET-10} & \multicolumn{6}{c|}{ACTNET-20} & \multirow{3}{*}{Params} \\[0.5pt]
    \cline{3-26}
    && \multicolumn{3}{c}{Recall} & \multicolumn{2}{c}{Rank} & \multirow{2}{*}{$BWF\downarrow$} & \multicolumn{3}{c}{Recall} & \multicolumn{2}{c}{Rank} & \multirow{2}{*}{$BWF\downarrow$} & \multicolumn{3}{c}{Recall} & \multicolumn{2}{c}{Rank} & \multirow{2}{*}{$BWF\downarrow$} & \multicolumn{3}{c}{Recall} & \multicolumn{2}{c}{Rank} & \multirow{2}{*}{$BWF\downarrow$} \\[0.5pt]
    \cline{3-7} \cline{9-13} \cline{15-19} \cline{21-25}
    && $@1\uparrow$ & $@5\uparrow$ & $@10\uparrow$ & $Med\downarrow$ & $Mean\downarrow$ & & $@1\uparrow$ & $@5\uparrow$ & $@10\uparrow$ & $Med\downarrow$ & $Mean\downarrow$ & & $@1\uparrow$ & $@5\uparrow$ & $@10\uparrow$ & $Med\downarrow$ & $Mean\downarrow$ & & $@1\uparrow$ & $@5\uparrow$ & $@10\uparrow$ & $Med\downarrow$ & $Mean\downarrow$ & \\[0.5pt]
    \hline
    \multirow{2}{*}{Zero-Shot CLIP \cite{radford2021CLIP}} & \multirow{2}{*}{$\times$} & 22.14 & 41.24 & 51.34 & 10.00 & 117.48 & 0.00 & 22.14 & 41.24 & 51.34 & 10.00 & 117.48 & 0.00 & 14.89 & 34.97 & 47.78 & 12.00 & 84.02 & 0.00 & 14.89 & 34.97 & 47.78 & 12.00 & 84.02 & 0.00 & \multirow{2}{*}{$0.00\text{M}$} \\
    && \textcolor{gray}{$\pm$0.00} & \textcolor{gray}{$\pm$0.00} & \textcolor{gray}{$\pm$0.00} & \textcolor{gray}{$\pm$0.00} & \textcolor{gray}{$\pm$0.00} & \textcolor{gray}{$\pm$0.00} & \textcolor{gray}{$\pm$0.00} & \textcolor{gray}{$\pm$0.00} & \textcolor{gray}{$\pm$0.00} & \textcolor{gray}{$\pm$0.00} & \textcolor{gray}{$\pm$0.00} & \textcolor{gray}{$\pm$0.00} & \textcolor{gray}{$\pm$0.00} & \textcolor{gray}{$\pm$0.00} & \textcolor{gray}{$\pm$0.00} & \textcolor{gray}{$\pm$0.00} & \textcolor{gray}{$\pm$0.00} & \textcolor{gray}{$\pm$0.00} & \textcolor{gray}{$\pm$0.00} & \textcolor{gray}{$\pm$0.00} & \textcolor{gray}{$\pm$0.00} & \textcolor{gray}{$\pm$0.00} & \textcolor{gray}{$\pm$0.00} & \textcolor{gray}{$\pm$0.00} & \\
    \hline
    \multirow{2}{*}{CLIP4Clip \cite{luo2022clip4clip}} & \multirow{2}{*}{$\times$} & 23.57 & 44.76 & 54.48 & 8.00 & 80.23 & 0.61 & 21.79 & 42.13 & 52.52 & 9.00 & 86.07 & 1.02 & 17.85 & 41.05 & \underline{54.88} & \underline{8.67} & \underline{54.97} & 0.75 & 17.07 & 39.96 & \underline{53.43} & 9.47 & 47.38 & 0.45 & \multirow{2}{*}{$151.28\text{M}$} \\
    && \textcolor{gray}{$\pm$0.37} & \textcolor{gray}{$\pm$0.24} & \textcolor{gray}{$\pm$0.61} & \textcolor{gray}{$\pm$0.00} & \textcolor{gray}{$\pm$1.32} & \textcolor{gray}{$\pm$0.37} & \textcolor{gray}{$\pm$0.20} & \textcolor{gray}{$\pm$0.30} & \textcolor{gray}{$\pm$0.35} & \textcolor{gray}{$\pm$0.00} & \textcolor{gray}{$\pm$0.47} & \textcolor{gray}{$\pm$0.46} & \textcolor{gray}{$\pm$0.06} & \textcolor{gray}{$\pm$0.88} & \textcolor{gray}{$\pm$0.33} & \textcolor{gray}{$\pm$0.58} & \textcolor{gray}{$\pm$0.82} & \textcolor{gray}{$\pm$0.08} & \textcolor{gray}{$\pm$0.09} & \textcolor{gray}{$\pm$0.57} & \textcolor{gray}{$\pm$0.45} & \textcolor{gray}{$\pm$0.16} & \textcolor{gray}{$\pm$0.41} & \textcolor{gray}{$\pm$0.22} & \\
    \multirow{2}{*}{X-Pool \cite{gorti2022XPool}} & \multirow{2}{*}{$\times$} & 19.60 & 39.80 & 49.49 & 11.00 & 94.89 & 0.28 & 15.98 & 34.21 & 44.34 & 15.33 & 105.97 & 1.37 & 17.99 & 39.81 & 52.38 & 9.67 & 60.49 & 0.37 & 16.57 & 39.83 & 51.82 & 10.22 & 55.00 & 0.31 & \multirow{2}{*}{$152.59\text{M}$} \\
    && \textcolor{gray}{$\pm$0.35} & \textcolor{gray}{$\pm$0.63} & \textcolor{gray}{$\pm$0.54} & \textcolor{gray}{$\pm$0.00} & \textcolor{gray}{$\pm$1.70} & \textcolor{gray}{$\pm$0.40} & \textcolor{gray}{$\pm$0.75} & \textcolor{gray}{$\pm$1.52} & \textcolor{gray}{$\pm$0.87} & \textcolor{gray}{$\pm$0.58} & \textcolor{gray}{$\pm$1.97} & \textcolor{gray}{$\pm$0.23} & \textcolor{gray}{$\pm$0.54} & \textcolor{gray}{$\pm$0.87} & \textcolor{gray}{$\pm$1.06} & \textcolor{gray}{$\pm$0.58} & \textcolor{gray}{$\pm$4.98} & \textcolor{gray}{$\pm$0.39} & \textcolor{gray}{$\pm$0.38} & \textcolor{gray}{$\pm$0.47} & \textcolor{gray}{$\pm$0.35} & \textcolor{gray}{$\pm$0.19} & \textcolor{gray}{$\pm$1.58} & \textcolor{gray}{$\pm$0.43} & \\
    \multirow{2}{*}{CLIP-ViP \cite{xue2023clipvip}} & \multirow{2}{*}{$\times$} & 21.56 & 44.19 & 53.43 & 8.00 & 86.71 & 0.49 & 19.74 & 41.25 & 50.61 & 10.00 & 93.95 & 0.73 & 17.01 & 38.73 & 52.01 & 9.67 & 59.66 & 0.56 & 16.02 & 37.29 & 50.92 & 10.58 & 48.78 & 0.73 & \multirow{2}{*}{$151.29\text{M}$} \\
    && \textcolor{gray}{$\pm$1.07} & \textcolor{gray}{$\pm$0.31} & \textcolor{gray}{$\pm$0.52} & \textcolor{gray}{$\pm$0.00} & \textcolor{gray}{$\pm$0.71} & \textcolor{gray}{$\pm$0.74} & \textcolor{gray}{$\pm$0.19} & \textcolor{gray}{$\pm$0.29} & \textcolor{gray}{$\pm$0.20} & \textcolor{gray}{$\pm$0.00} & \textcolor{gray}{$\pm$1.19} & \textcolor{gray}{$\pm$0.31} & \textcolor{gray}{$\pm$0.53} & \textcolor{gray}{$\pm$0.82} & \textcolor{gray}{$\pm$0.75} & \textcolor{gray}{$\pm$0.58} & \textcolor{gray}{$\pm$1.50} & \textcolor{gray}{$\pm$0.21} & \textcolor{gray}{$\pm$0.19} & \textcolor{gray}{$\pm$0.21} & \textcolor{gray}{$\pm$0.71} & \textcolor{gray}{$\pm$0.38} & \textcolor{gray}{$\pm$1.16} & \textcolor{gray}{$\pm$0.13} & \\
    \hline
    \multirow{2}{*}{LwF \cite{li2017LwF}}  & \multirow{2}{*}{$\checkmark$} & 23.85 & 45.30 & \underline{55.68} & \underline{7.33} & 76.46 & 1.68 & 22.06 & 42.77 & 52.69 & 9.00 & 85.27 & 1.65 & 17.56 & 40.18 & 53.67 & 9.00 & 55.33 & 0.63 & 16.36 & \underline{40.14} & 53.29 & 9.44 & \underline{45.94} & 0.93 & \multirow{2}{*}{$151.28\text{M}$} \\
    && \textcolor{gray}{$\pm$0.09} & \textcolor{gray}{$\pm$0.26} & \textcolor{gray}{$\pm$0.32} & \textcolor{gray}{$\pm$0.58} & \textcolor{gray}{$\pm$0.44} & \textcolor{gray}{$\pm$0.59} & \textcolor{gray}{$\pm$0.44} & \textcolor{gray}{$\pm$1.33} & \textcolor{gray}{$\pm$0.91} & \textcolor{gray}{$\pm$1.00} & \textcolor{gray}{$\pm$3.99} & \textcolor{gray}{$\pm$0.72} & \textcolor{gray}{$\pm$0.12} & \textcolor{gray}{$\pm$0.20} & \textcolor{gray}{$\pm$0.41} & \textcolor{gray}{$\pm$0.00} & \textcolor{gray}{$\pm$1.86} & \textcolor{gray}{$\pm$0.42} & \textcolor{gray}{$\pm$0.31} & \textcolor{gray}{$\pm$0.31} & \textcolor{gray}{$\pm$0.46} & \textcolor{gray}{$\pm$0.25} & \textcolor{gray}{$\pm$2.22} & \textcolor{gray}{$\pm$0.14} & \\
    \multirow{2}{*}{VR-LwF \cite{ding2022VR_LwF}} & \multirow{2}{*}{$\checkmark$} & \underline{24.49} & \underline{45.59} & 55.45 & \underline{7.33} & \underline{74.89} & 1.22 & 22.39 & \underline{43.27} & \underline{53.33} & \underline{8.67} & \underline{82.04} & 1.44 & \underline{18.08} & \textbf{41.44} & \textbf{54.98} & \textbf{8.50} & \textbf{53.28} & 0.68 & \underline{17.21} & \textbf{40.96} & \textbf{54.18} & \textbf{9.00} & \textbf{44.45} & 0.58 & \multirow{2}{*}{$151.28\text{M}$} \\
    && \textcolor{gray}{$\pm$0.20} & \textcolor{gray}{$\pm$1.14} & \textcolor{gray}{$\pm$0.89} & \textcolor{gray}{$\pm$0.58} & \textcolor{gray}{$\pm$2.56} & \textcolor{gray}{$\pm$0.46} & \textcolor{gray}{$\pm$0.43} & \textcolor{gray}{$\pm$0.43} & \textcolor{gray}{$\pm$0.96} & \textcolor{gray}{$\pm$0.58} & \textcolor{gray}{$\pm$2.16} & \textcolor{gray}{$\pm$0.16} & \textcolor{gray}{$\pm$0.55} & \textcolor{gray}{$\pm$0.36} & \textcolor{gray}{$\pm$0.45} & \textcolor{gray}{$\pm$0.50} & \textcolor{gray}{$\pm$2.39} & \textcolor{gray}{$\pm$0.47} & \textcolor{gray}{$\pm$0.36} & \textcolor{gray}{$\pm$0.26} & \textcolor{gray}{$\pm$0.15} & \textcolor{gray}{$\pm$0.13} & \textcolor{gray}{$\pm$0.70} & \textcolor{gray}{$\pm$0.13} & \\
    \multirow{2}{*}{ZSCL \cite{zheng2023ZSCL}} & \multirow{2}{*}{$\checkmark$} & 23.99 & 45.15 & 54.77 & 8.00 & 79.69 & \underline{0.10} & 21.47 & 41.61 & 52.05 & 9.33 & 88.45 & 0.91 & 17.67 & 41.05 & 54.05 & 9.00 & 55.74 & 0.35 & 16.83 & 38.90 & 52.07 & \underline{9.33} & 65.03 & 0.70 & \multirow{2}{*}{$151.28\text{M}$} \\
    && \textcolor{gray}{$\pm$0.44} & \textcolor{gray}{$\pm$0.33} & \textcolor{gray}{$\pm$0.24} & \textcolor{gray}{$\pm$0.00} & \textcolor{gray}{$\pm$0.95} & \textcolor{gray}{$\pm$0.78} & \textcolor{gray}{$\pm$0.77} & \textcolor{gray}{$\pm$0.87} & \textcolor{gray}{$\pm$0.92} & \textcolor{gray}{$\pm$0.58} & \textcolor{gray}{$\pm$6.38} & \textcolor{gray}{$\pm$0.33} & \textcolor{gray}{$\pm$0.55} & \textcolor{gray}{$\pm$0.47} & \textcolor{gray}{$\pm$0.15} & \textcolor{gray}{$\pm$0.00} & \textcolor{gray}{$\pm$0.19} & \textcolor{gray}{$\pm$0.45} & \textcolor{gray}{$\pm$0.17} & \textcolor{gray}{$\pm$0.55} & \textcolor{gray}{$\pm$0.31} & \textcolor{gray}{$\pm$0.58} & \textcolor{gray}{$\pm$1.59} & \textcolor{gray}{$\pm$0.08} & \\
    \multirow{2}{*}{MoE-Adapter \cite{yu2024MoEAdapter}} & \multirow{2}{*}{$\times$} & 22.92 & 42.76 & 52.11 & 9.00 & 105.70 & 0.14 & \underline{22.70} & 41.96 & 51.82 & 9.00 & 112.86 & \underline{0.01} & 16.63 & 37.29 & 50.36 & 10.33 & 70.49 & \textbf{-0.15} & 15.77 & 36.27 & 49.32 & 11.00 & 77.65 & \textbf{-0.01} & \multirow{2}{*}{$\underline{59.8\text{M}}$} \\
    && \textcolor{gray}{$\pm$0.09} & \textcolor{gray}{$\pm$0.24} & \textcolor{gray}{$\pm$0.09} & \textcolor{gray}{$\pm$0.00} & \textcolor{gray}{$\pm$2.66} & \textcolor{gray}{$\pm$0.11} & \textcolor{gray}{$\pm$0.14} & \textcolor{gray}{$\pm$0.16} & \textcolor{gray}{$\pm$0.10} & \textcolor{gray}{$\pm$0.00} & \textcolor{gray}{$\pm$0.34} & \textcolor{gray}{$\pm$0.05} & \textcolor{gray}{$\pm$0.55} & \textcolor{gray}{$\pm$0.48} & \textcolor{gray}{$\pm$0.89} & \textcolor{gray}{$\pm$0.58} & \textcolor{gray}{$\pm$3.70} & \textcolor{gray}{$\pm$0.08} & \textcolor{gray}{$\pm$0.11} & \textcolor{gray}{$\pm$0.14} & \textcolor{gray}{$\pm$0.24} & \textcolor{gray}{$\pm$0.00} & \textcolor{gray}{$\pm$1.23} & \textcolor{gray}{$\pm$0.27} & \\
    \hline
    \multirow{2}{*}{TVR \cite{xue2023clipvip} + CL \cite{ding2022VR_LwF}} & \multirow{2}{*}{$\checkmark$} & 22.47 & 43.71 & 53.59 & 8.00 & 82.97 & 0.46 & 21.28 & 42.28 & 51.82 & 9.67 & 89.22 & 1.28 & 16.88 & 38.87 & 51.82 & 9.67 & 61.16 & 0.44 & 16.37 & 37.78 & 50.51 & 10.33 & 66.80 & 0.76 & \multirow{2}{*}{$151.29\text{M}$} \\
    && \textcolor{gray}{$\pm$0.55} & \textcolor{gray}{$\pm$0.42} & \textcolor{gray}{$\pm$0.28} & \textcolor{gray}{$\pm$0.00} & \textcolor{gray}{$\pm$1.31} & \textcolor{gray}{$\pm$0.17} & \textcolor{gray}{$\pm$0.66} & \textcolor{gray}{$\pm$1.10} & \textcolor{gray}{$\pm$0.63} & \textcolor{gray}{$\pm$0.58} & \textcolor{gray}{$\pm$3.44} & \textcolor{gray}{$\pm$0.73} & \textcolor{gray}{$\pm$0.62} & \textcolor{gray}{$\pm$0.70} & \textcolor{gray}{$\pm$1.45} & \textcolor{gray}{$\pm$0.58} & \textcolor{gray}{$\pm$4.34} & \textcolor{gray}{$\pm$0.27} & \textcolor{gray}{$\pm$0.27} & \textcolor{gray}{$\pm$1.08} & \textcolor{gray}{$\pm$1.11} & \textcolor{gray}{$\pm$0.58} & \textcolor{gray}{$\pm$5.46} & \textcolor{gray}{$\pm$0.26} & \\
    \hline
     \rowcolor{lightpurple}&  & \textbf{25.87} & \textbf{45.91} & \textbf{56.03} & \textbf{7.00} & \textbf{74.70} & \textbf{-0.45} & \textbf{25.16} & \textbf{45.53} & \textbf{55.10} & \textbf{7.33} & \textbf{77.79} & \textbf{-0.70} & \textbf{18.21} & \underline{40.45} & 53.94 & 9.00 & 56.14 & \underline{-0.01} & \textbf{17.71} & 39.40 & 52.76 & \textbf{9.00} & 62.22 & \underline{0.04} &  \\
    \rowcolor{lightpurple}
    \multirow{-2}{*}{\textbf{StableFusion} } & \multirow{-2}{*}{$\times$}
    &  \textcolor{gray}{$\pm$0.33} & \textcolor{gray}{$\pm$0.03} & \textcolor{gray}{$\pm$0.64} & \textcolor{gray}{$\pm$0.00} & \textcolor{gray}{$\pm$1.06} & \textcolor{gray}{$\pm$0.22} & \textcolor{gray}{$\pm$0.14} & \textcolor{gray}{$\pm$0.42} & \textcolor{gray}{$\pm$0.07} & \textcolor{gray}{$\pm$0.58} & \textcolor{gray}{$\pm$0.57} & \textcolor{gray}{$\pm$0.25} & \textcolor{gray}{$\pm$0.33} & \textcolor{gray}{$\pm$0.29} & \textcolor{gray}{$\pm$0.34} & \textcolor{gray}{$\pm$0.00} & \textcolor{gray}{$\pm$0.63} & \textcolor{gray}{$\pm$0.40} & \textcolor{gray}{$\pm$0.25} & \textcolor{gray}{$\pm$0.08} & \textcolor{gray}{$\pm$0.15} & \textcolor{gray}{$\pm$0.00} & \textcolor{gray}{$\pm$1.99} & \textcolor{gray}{$\pm$0.06} & \multirow{-2}{*}{\textbf{$46.8\text{M}$}} \\
    \hline
    \multirow{2}{*}{\textit{Upper Bound}} & \multirow{2}{*}{$\times$} & 25.86 & 48.34 & 58.96 & 6.00 & 65.10 & -0.17 & 25.80 & 48.54 & 58.84 & 6.00 & 65.13 & 0.23 & 19.66 & 44.57 & 59.18 & 7.00 & 37.16 & 0.27 & 19.78 & 44.63 & 58.96 & 7.00 & 36.18 & 0.16 & \multirow{2}{*}{$151.28\text{M}$} \\
    && \textcolor{gray}{$\pm$0.67} & \textcolor{gray}{$\pm$0.25} & \textcolor{gray}{$\pm$0.12} & \textcolor{gray}{$\pm$0.00} & \textcolor{gray}{$\pm$1.09} & \textcolor{gray}{$\pm$1.21} & \textcolor{gray}{$\pm$0.45} & \textcolor{gray}{$\pm$1.04} & \textcolor{gray}{$\pm$0.65} & \textcolor{gray}{$\pm$0.00} & \textcolor{gray}{$\pm$1.79} & \textcolor{gray}{$\pm$0.44} & \textcolor{gray}{$\pm$0.08} & \textcolor{gray}{$\pm$0.17} & \textcolor{gray}{$\pm$0.23} & \textcolor{gray}{$\pm$0.00} & \textcolor{gray}{$\pm$0.18} & \textcolor{gray}{$\pm$0.10} & \textcolor{gray}{$\pm$0.17} & \textcolor{gray}{$\pm$0.31} & \textcolor{gray}{$\pm$0.24} & \textcolor{gray}{$\pm$0.00} & \textcolor{gray}{$\pm$0.76} & \textcolor{gray}{$\pm$0.31} & \\
    \hline
    \end{tabular}}
    \label{tab:Overview}
    \vspace{-0.2cm}
\end{table*}

\vspace{-0.2cm}
\subsection{Evaluation Metrics}
\label{sec:eval}
Following standard TVR evaluation \cite{luo2022clip4clip, gorti2022XPool}, we measure how well the model performs in retrieval across all learned tasks by reporting Recall@1 (\textbf{R@1}), Recall@5 (\textbf{R@5}), Recall@10 (\textbf{R@10}), Median Rank (\textbf{MedR}) and Mean Rank (\textbf{MeanR}). When evaluating at task $t$, we test the model on queries from both current and all previous tasks $\mQ_{[1:t]}$. The search space consists of all videos from the video feature database $\mV_{[1:t]}$, where $\mV_{[1:t-1]}$ are extracted and stored using the models learned in previous tasks. Videos are ranked by cosine similarity between query and video features. 

To measure how learning new tasks affects the model's performance on previous tasks, we evaluate \textbf{Backward Forgetting (BWF)}.  When testing on task $t$, we measure the performance drop of each previous task $i$ (where $i < t$) by comparing its current performance with its performance right after the task was initially learned. Formally, the BWF at task $n$ is defined as   $\text{BWF}_{t} = \frac{1}{t-1} \sum_{i=1}^{t-1} (\text{R}_{i,i} - \text{R}_{t,i}) $,
where $\text{R}_{i,i}$ represents the R@1 on the queries of task $i$ after training on task $i$, and $\text{R}_{t,i}$ measures the R@1 on the queries of task $i$ after training on task $t$.

\subsection{Benchmarking Baselines}
\label{sec:baselines}

\subsubsection{\textbf{TVR Methods}}
Building upon CLIP's \cite{radford2021CLIP} vision-language embedding capabilities, recent TVR models aim to transfer knowledge from the image-text domain to the video-text domain.

\noindent \textbf{Parameter-free Methods.} \textbf{CLIP4Clip} \cite{luo2022clip4clip} introduces the first parameter-free TVR model (AvgPool) based on CLIP, where video features are obtained by decomposing videos into image sequences and applying average pooling on a sequence of frame features. \textbf{Upper Bound} represents the performance ceiling based on CLIP4Clip \cite{luo2022clip4clip}, where the model is trained with full access to the combined data from all tasks simultaneously in an i.i.d. manner.

\noindent \textbf{Architecture-enhanced Methods.} These approaches focus on architectural modifications on CLIP to enable video temporal feature learning capabilities, thereby enhancing video-text representations. \textbf{X-Pool} \cite{gorti2022XPool} introduce a Transformer block on top of the vision-language embedding space that learns the relevance between queries and individual video frames. \textbf{CLIP-ViP} \cite{xue2023clipvip} introduces a lightweight architectural modification that augments CLIP's temporal learning capacity through the incorporation of temporal embeddings and video proxy vectors.

\vspace{-0.2cm}
\subsubsection{\textbf{Vision-Language CL Methods}}
Our benchmark includes CL approaches that are established with CLIP. 

\noindent \textbf{Knowledge Distillation-based Methods.} \textbf{LwF} \cite{li2017LwF} utilizes the model from the previous task as a teacher model, employing distillation loss to maintain feature preservation for older tasks during current task learning.  \textbf{VR-LwF} \cite{ding2022VR_LwF} extends the LwF framework with a focus on maintaining CLIP's inherent text-vision alignment capabilities. Through random sampling from CLIP's vocabulary set to create a replayed vocabulary set during training, it achieves better mitigation of catastrophic forgetting. \textbf{ZSCL} \cite{zheng2023ZSCL} employs a reference dataset to preserve CLIP's original image-text alignment.

\noindent \textbf{Dynamic Network Methods.} \textbf{MoE-Adapter} \cite{yu2024MoEAdapter} introduces a MoE-structured adapters to preserve the zero-shot capabilities of vision-language models. Through routing and gating operations, MoE dynamically selects appropriate experts.

To eliminate confounding factors from TVR model architectures, our benchmark implements these CL baselines on the \textbf{CLIP4Clip}, maintaining maximum consistency with their original deployment methodologies. In addition, we also consider the combination between \textbf{VR-LwF} and \textbf{CLIP-ViP}.

\begin{figure*}
    \centering
    \includegraphics[width=1\linewidth]{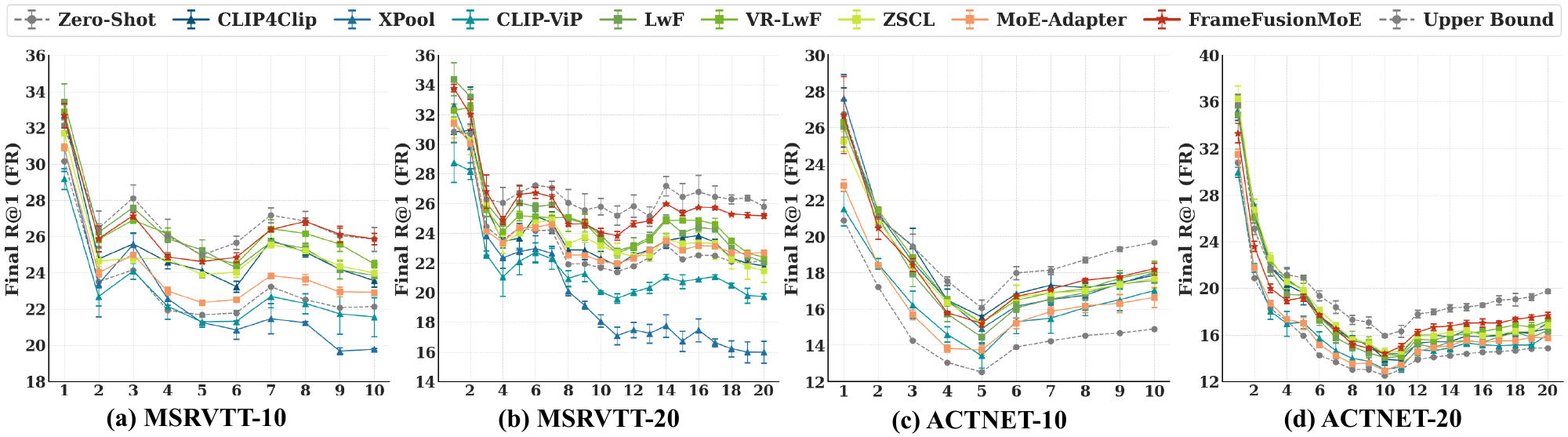}
    \vspace{-0.6cm}
    \Description{}
    \caption{Comparative analysis of Final R@1 (FR) performance for different CTVR configurations across all tasks.}
    \label{fig:CL_performance}
    \vspace{-0.15cm}
\end{figure*}

\vspace{-0.3cm}
\subsection{\textbf{Implementation Details}}
\label{sec:implementation}
For all experiments, we utilize CLIP-B/32 as the pre-trained model with 20 training epochs per task and 16 videos per class. All methods employ a cosine learning rate scheduler. For baseline-specific parameters, CLIP-ViP \cite{xue2023clipvip} is configured with 4 video proxies and methods involving knowledge distillation \cite{li2017LwF, ding2022VR_LwF, zheng2023ZSCL} are implemented with a temperature of $\tau = 2.0$. All methods are evaluated with mean and standard deviation across three different random seeds. For experiments on MSRVTT and ACTNET datasets, we sample 12 and 24 frames per video with batch sizes of 8 and 16, respectively. For our method, we employ learning rates of $4 \times 10^{-6}$ and $6 \times 10^{-6}$, respectively. The frame fusion adapter consists of 10 layers for MSRVTT and 12 layers for ACTNET, with each TAME layer containing 10 experts for MSRVTT and 5 experts for ACTNET, and a loss scale of $\beta = 0.6$.



\vspace{-0.2cm}
\section{Results and Analysis}
\label{sec:exp}

To comprehensively evaluate the effectiveness of our proposed method, we conduct extensive experiments to answer the following research questions: (1) \textbf{RQ1}: How does FrameFusionMoE perform compared to existing TVR and CL baselines in terms of effectiveness on CTVR? (2) \textbf{RQ2}: How do the main architectural components of FrameFusionMoE contribute to its overall performance? (3) \textbf{RQ3}: How do key hyper-parameters of FrameFusionMoE affect its continual retrieval performance?

\vspace{-0.2cm}
\subsection{Overall Performance (RQ1)}
Table \textcolor{red}{\ref{tab:Overview}} and Figure \textcolor{red}{\ref{fig:CL_performance}} present a comparative analysis of the CTVR performance and backward forgetting between FrameFusionMoE and the baseline approaches across two datasets with four CL configurations. We observe the following key observations:

\vspace{5pt}
\noindent\textbf{Effectiveness of FrameFusionMoE.}
Across all CTVR configurations, FrameFusionMoE surpasses all baseline models, showing R@1 improvements of +1.38, +2.46, +0.13, and +0.50 over the second-best model on all datasets. Figure \textcolor{red}{\ref{fig:CL_performance}} demonstrates that our model achieves higher R@1 than all baselines across different CTVR configurations and tasks. 
Notably, our method tends to establish higher performance gain when learning through more tasks, which demonstrates the continual learning capability in long-term tasks.


Our method achieves near-zero BWF scores (-0.70 to +0.04) across multiple CL configurations, indicating minimal catastrophic forgetting. Notably, in several configurations, we observe negative BWF scores, suggesting that learning new tasks enhances the model's performance on previous tasks. FrameFusionMoE achieves these improvements while maintaining computational efficiency. Compared to standard CL baselines, FrameFusionMoE involves only 30.94\% of the trainable parameters. Compared to the parameter-efficient MoE-Adapter \cite{yu2024MoEAdapter}, we use 70.26\% of its parameters. 

\vspace{5pt}
\noindent {\textbf{Model Plasticity Loss of TVR Methods.}}
TVR baselines exhibit unexpected performance patterns in CTVR benchmark compared to the one-off learning scenario. Our experiments show that the CLIP4clip, which preserves CLIP's original architecture, consistently outperforms more advanced TVR models across various CL configurations. This counter-intuitive result is particularly pronounced in MSRVTT. As shown in Figure \textcolor{red}{\ref{fig:plasticity_loss}}, carefully designed architectures like XPool and CLIP-ViP significantly degrade CLIP's original generalization capability after learning the first few tasks, making it difficult to adapt to subsequent tasks with limited training data. This loss of model plasticity cascades through the continual learning process.

\vspace{5pt}
\noindent  {\textbf{Catastrophic Forgetting on CL Baselines.}}
Our experimental results indicate that these Continual Learning (CL) baselines demonstrate less significant improvements in TVR tasks compared to their established performance gains in image recognition \cite{li2017LwF, ding2022VR_LwF, zheng2023ZSCL, yu2024MoEAdapter}. This performance disparity stems from fundamental differences in objectives: retrieval tasks are particularly susceptible to Representation Shift, which significantly impacts performance in CTVR scenarios. As demonstrated in Figure \textcolor{red}{\ref{fig:shift}}, fine-tuning on new tasks causes text embedding representation shift, leading to a progressive misalignment between video and query representations in the shared embedding space. This explains why CL baselines such as LwF experience a performance drop of 2.1 as the number of tasks increases from 10 to 20 on the MSRVTT.

\begin{table}[tp]\footnotesize
    \centering
    \definecolor{red}{RGB}{255,150,150}
    \definecolor{lightgreen}{RGB}{150,200,150}
    \caption{Ablation study of the contribution of individual components in our proposed method. We report the performance impact of removing Frame Fusion Adapter (FFA), Task-Aware Mixture-of-Experts (TAME), Task-Specific Prototype (TP) modules and Cross-Task Loss ($\mathcal{L}_{CT}$). The $\Delta$ represent performance degradation relative to the complete model highlighted in \textcolor{red}{red}.}
    \vspace{-0.2cm}
    \setlength{\tabcolsep}{0.1pt}
    \renewcommand{\arraystretch}{1.2}
    \scalebox{0.95}{\begin{tabular}{lccccccccccccc}
    \hline
    \multirow{2}{*}{\textbf{Model}} & \multicolumn{6}{c}{\textbf{MSRVTT}} & \multicolumn{6}{c}{\textbf{ActivityNet}} \\
    \cline{2-13}
    & \multicolumn{3}{c}{10 Task} & \multicolumn{3}{c}{20 Task} & \multicolumn{3}{c}{10 Task} & \multicolumn{3}{c}{20 Task} \\
    \cline{2-13}
    & R@1 $\uparrow$ & $\Delta$ & BWF $\downarrow$ & R@1 $\uparrow$ & $\Delta$ & BWF $\downarrow$ & R@1 $\uparrow$ & $\Delta$ & BWF $\downarrow$ & R@1 $\uparrow$ & $\Delta$ & BWF $\downarrow$ \\
    \hline
    Ours & 26.25 &  & -0.69 & 25.32 &  & -0.86 & 18.49 &  & -0.26 & 17.92 &  & 0.08 \\
    \hline
    w/o FFA & 17.59 & \textcolor{red}{-8.66} & 0.13 & 20.60 & \textcolor{red}{-4.72} & -0.21 & 16.84 & \textcolor{red}{-1.65} & -0.28 & 16.42 & \textcolor{red}{-1.50} & -0.44 \\
    w/o TAME & 25.22 & \textcolor{red}{-1.03} & -0.04 & 24.58 & \textcolor{red}{-0.74} & 0.62 & 16.13 & \textcolor{red}{-2.36} & 0.27 & 15.76 & \textcolor{red}{-2.16} & -0.43 \\
    w/o TP & 25.92 & \textcolor{red}{-0.33} & -0.98 & 24.78 & \textcolor{red}{-0.54} & -0.80 & 17.72 & \textcolor{red}{-0.77} & 0.39 & 17.90 & \textcolor{red}{-0.02} & -0.29 \\
    w/o $\mathcal{L}_{CT}$ & 25.15 & \textcolor{red}{-1.10} & -0.09 & 24.21 & \textcolor{red}{-1.11} & -0.16 & 17.70 & \textcolor{red}{-0.79} & 0.42 & 17.27 & \textcolor{red}{-0.65} & 0.38 \\
    \hline
    Baseline \cite{luo2022clip4clip} & 23.18 & \textcolor{red}{-3.07} & 0.96 & 21.87 & \textcolor{red}{-3.45} & 1.15 & 17.92 & \textcolor{red}{-0.57} & 0.76 & 17.11 & \textcolor{red}{-0.81} & 0.46 \\
    \hline
    \end{tabular}}
    \label{tab:EffectivenessOfModules}
    \vspace{-0.8em}
\end{table}

\vspace{-0.2cm}
\subsection{Effectiveness of Components (RQ2).}
To thoroughly validate the effectiveness of our proposed method, we conduct an ablation study on the key components of our architecture.
In Table \textcolor{red}{\ref{tab:EffectivenessOfModules}}, we respectively remove individual components to measure their contribution to the entire framework.
\noindent \textbf{FrameFusion Adapter (FFA):} The FFA module enables efficient temporal video learning, and its removal causes model plasticity loss.
\noindent \textbf{Task-Aware Mixture-of-Experts (TAME):} The TAME module maintains query-feature alignment across tasks, and its removal leads to catastrophic forgetting in the shared feature space.
\noindent \textbf{Task-Prototype (TP):} The TP module learns task-specific features to guide TAME's routing, and its removal impairs the model's task discrimination capability.
\noindent \textbf{Cross-Task Loss ($\mathcal{L}_{CT}$)} maximizes the distance between current query-video features and cached video representations in the feature space, and its removal causes feature overlap across different tasks.

\noindent \textbf{Effectiveness of Individual Components.} Our ablation studies demonstrate that every component contributes to model performance, as removing any component results in significant performance degradation across different tasks. The ablation study demonstrates the crucial role of each proposed component. First, removing \textbf{FFA} causes the most severe performance drop, underscoring its fundamental importance in maintaining model plasticity. Second, the absence of \textbf{TAME} leads to increased backward forgetting, validating its effectiveness in mitigating catastrophic forgetting. The \textbf{TP} module, built upon TAME, further enhances retrieval performance through its prototype selection mechanism, as evidenced by the performance gap when removing TP alone. The \textbf{$\mathcal{L}_{CT}$} effectively captures task-wise embedding overlaps, demonstrated by consistent R@1 decreases across datasets when removed. 

\begin{figure}[t]
    \centering
    \includegraphics[width=1\linewidth]{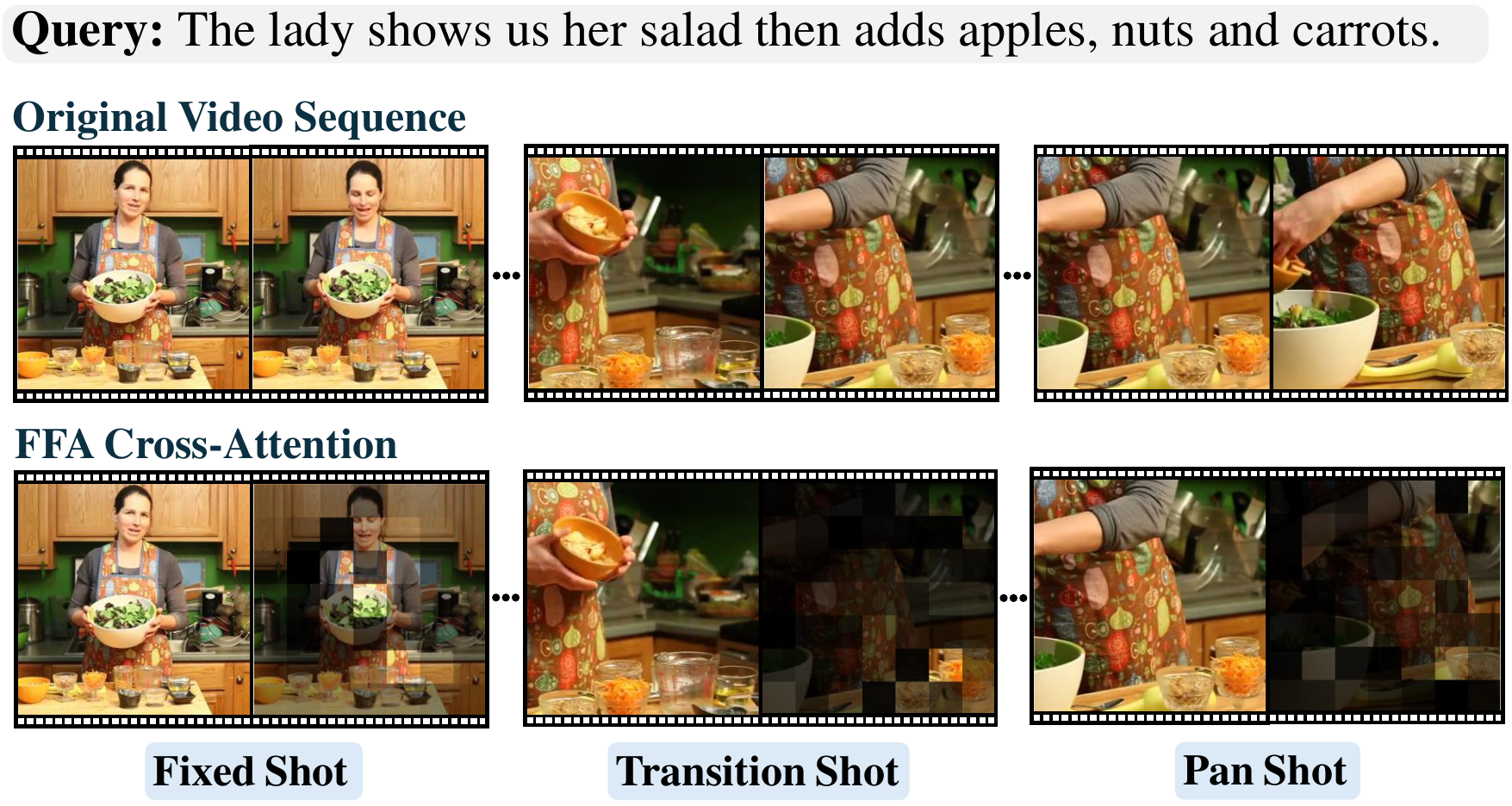}
    \vspace{-0.5cm}
    \Description{}
    \caption{Visualization of FFA temporal cross-attention mechanism across different shot types in video sequences.}
    \label{fig:FFA_vis}
    \vspace{-0.5cm}
\end{figure}

\noindent \textbf{Analysis of FFA Temporal Cross-attention Mechanism.} Figure \textcolor{red}{\ref{fig:FFA_vis}} visualizes how FFA cross-attention works in video sequences across different types of camera shots. In fixed shot (left), where frames remain mostly static, FFA maintains balanced attention distribution despite changes in object actions, effectively capturing inter-frame relationships. In transition shot (middle), where scene composition changes significantly between frames, FFA's attention mechanism focuses on identifying visual consistencies (\textit{e.g.,} carrot bowl) to maintain continuity of shared objects and backgrounds. In pan shot (right), where the camera moves upward-right, FFA maintains attention between corresponding regions across frames, with stronger attention weights on overlapping elements between consecutive frames.

\vspace{-0.2cm}
\subsection{Hyper-parameter Study (RQ3)}
\begin{figure}[t]
    \centering
    \includegraphics[width=1\linewidth]{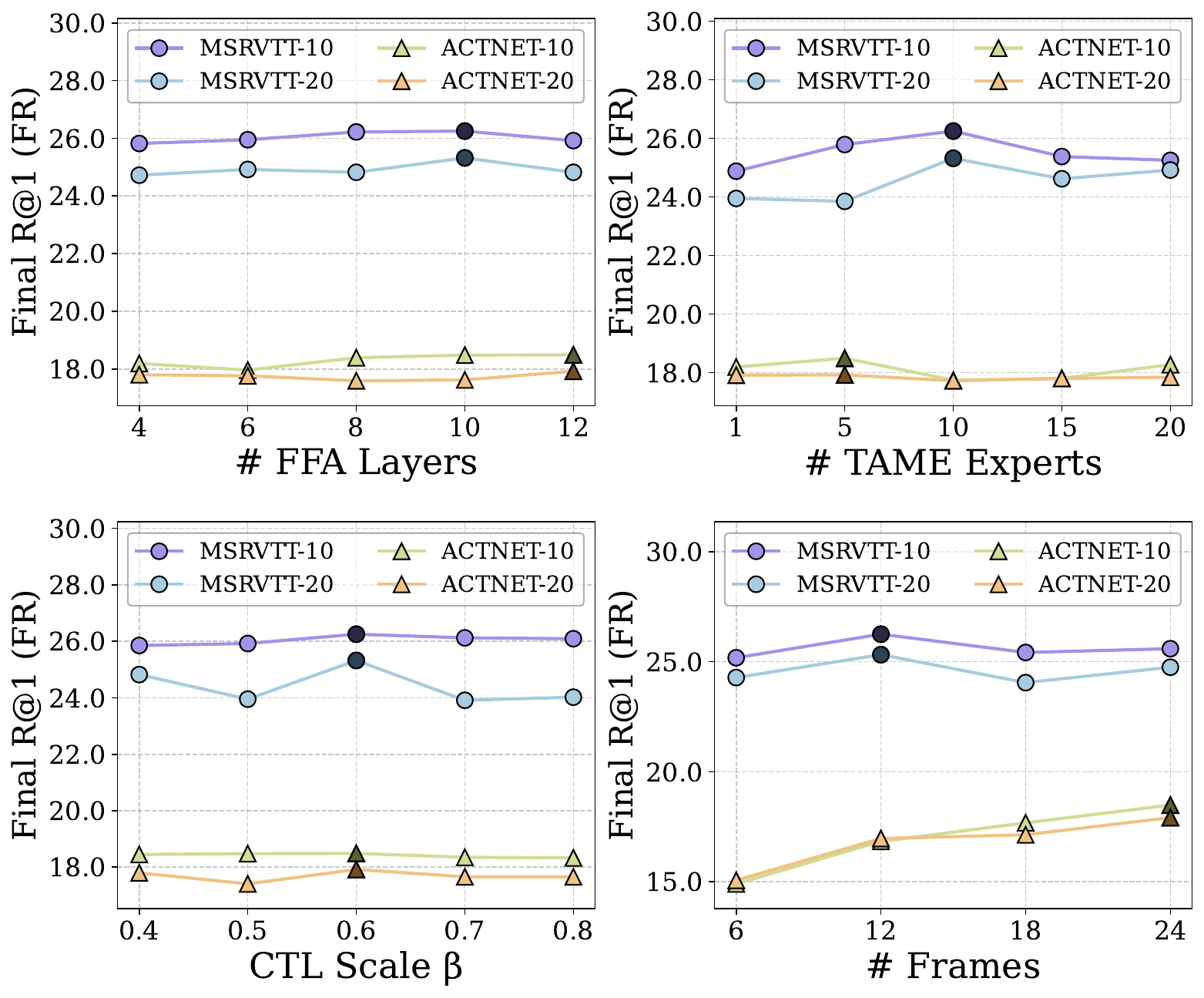}
    \vspace{-0.6cm}
    \Description{}
    \caption{Hyper-parameter sensitivity on two datasets with 10 and 20 tasks respectively.}
    \label{fig:hyper_tuning}
    \vspace{-0.6cm}
\end{figure}

We conduct further analysis to understand how key hyper-parameters impact our model's performance across all datasets in Figure \textcolor{red}{\ref{fig:hyper_tuning}}. \\
\noindent \textbf{Number of FFA adapters:} We investigate the impact of FFA adapters by implementing them starting from the shallowest layers of the CLIP Transformer blocks. As illustrated in Figure \textcolor{red}{\ref{fig:hyper_tuning}}, we experiment with 4 to 12 (covering all blocks) FFA adapters across various CL settings. The results demonstrate consistent performance trends within each dataset, with 10-12 FFA adapters emerging as the optimal range, highlighting the FFA adapter's effectiveness in learning video temporal information. \\
\noindent \textbf{Number of experts used in TAME:} We analyze the effect of varying the number of experts in TAME from 1 to 20. Experimental results demonstrate that model performance does not scale linearly with the number of experts. Our analysis reveals that configurations with 5 and 10 experts achieve the optimal trade-off between model effectiveness and computational overhead. \\
\noindent\textbf{Scales of CTL ($\mathcal{L}_{CT}$):} We investigate the impact of scaling the CTL ($\mathcal{L}_{CT}$) component in the overall objective function. As shown in Figure \textcolor{red}{\ref{fig:hyper_tuning}}, adjusting the scaling coefficient of the $\mathcal{L}_{CT}$ demonstrated marginal impact on the model's final R@1. Through evaluation of different scaling factors, we find that setting the $\mathcal{L}_{CT}$ coefficient to 0.6 achieved optimal performance by effectively balancing the contributions of in-task and cross-task contrastive objectives. \\
\noindent\textbf{Number of sampled frames:} We conduct experiments on the number of sampled frames per video during both training and inference phases. As shown in Figure \textcolor{red}{\ref{fig:hyper_tuning}}, in MSRVTT dataset, the model performance does not consistently improve with increased frames. Specifically, undersampling at 6 frames likely leads to performance degradation due to loss of critical temporal information, while excessive frame sampling introduces noise that adversely affects the quality of video representations. In contrast, for ACTNET dataset, as specified in section \ref{sec:datasets}, since its videos are approximately five times longer than those in MSRVTT, increasing the number of sampled frames leads to performance improvements by better capturing the extended temporal dynamics.

\section{Conclusion}
In this paper, we introduce the novel research problem of Continual Text-to-Video Retrieval (CTVR), addressing the challenges posed by the dynamic and evolving nature of video content. To tackle the limitations of existing TVR and CL approaches, we propose FrameFusionMoE to maintain model plasticity while mitigating catastrophic forgetting. FrameFusionMoE incorporates the Frame Fusion Adapter (FFA) for capturing video temporal dynamics, and the Task-Aware Mixture-of-Experts (TAME) for ensuring consistent query-video alignment across tasks through task-aware routing.
Comprehensive experiments demonstrate that FrameFusionMoE consistently outperforms existing methods. 
By establishing a benchmark for CTVR and providing a detailed analysis of current state-of-the-art methods, our work paves the way for future research in this emerging field. 


\begin{acks}
This work is supported by the Australian Research Council under the streams of Discovery Project (No. DP240101814), Discovery Early Career Research Award (No. DE230101033), Linkage Project (No. LP230200892), and Centre of Excellence (No. CE200100025).
\end{acks}

\balance
\bibliographystyle{ACM-Reference-Format}
\bibliography{references}


\begin{thebibliography}{79}


\ifx \showCODEN    \undefined \def \showCODEN     #1{\unskip}     \fi
\ifx \showDOI      \undefined \def \showDOI       #1{#1}\fi
\ifx \showISBNx    \undefined \def \showISBNx     #1{\unskip}     \fi
\ifx \showISBNxiii \undefined \def \showISBNxiii  #1{\unskip}     \fi
\ifx \showISSN     \undefined \def \showISSN      #1{\unskip}     \fi
\ifx \showLCCN     \undefined \def \showLCCN      #1{\unskip}     \fi
\ifx \shownote     \undefined \def \shownote      #1{#1}          \fi
\ifx \showarticletitle \undefined \def \showarticletitle #1{#1}   \fi
\ifx \showURL      \undefined \def \showURL       {\relax}        \fi
\providecommand\bibfield[2]{#2}
\providecommand\bibinfo[2]{#2}
\providecommand\natexlab[1]{#1}
\providecommand\showeprint[2][]{arXiv:#2}

\bibitem[Aljundi et~al\mbox{.}(2018)]%
        {aljundi2018MAS}
\bibfield{author}{\bibinfo{person}{Rahaf Aljundi}, \bibinfo{person}{Francesca Babiloni}, \bibinfo{person}{Mohamed Elhoseiny}, \bibinfo{person}{Marcus Rohrbach}, {and} \bibinfo{person}{Tinne Tuytelaars}.} \bibinfo{year}{2018}\natexlab{}.
\newblock \showarticletitle{Memory aware synapses: Learning what (not) to forget}. In \bibinfo{booktitle}{\emph{Proceedings of the European conference on computer vision (ECCV)}}. \bibinfo{pages}{139--154}.
\newblock


\bibitem[Caba~Heilbron et~al\mbox{.}(2015)]%
        {caba2015activitynet}
\bibfield{author}{\bibinfo{person}{Fabian Caba~Heilbron}, \bibinfo{person}{Victor Escorcia}, \bibinfo{person}{Bernard Ghanem}, {and} \bibinfo{person}{Juan Carlos~Niebles}.} \bibinfo{year}{2015}\natexlab{}.
\newblock \showarticletitle{Activitynet: A large-scale video benchmark for human activity understanding}. In \bibinfo{booktitle}{\emph{Proceedings of the ieee conference on computer vision and pattern recognition}}. \bibinfo{pages}{961--970}.
\newblock


\bibitem[Chen et~al\mbox{.}(2025)]%
        {chen2025promptfusion_mm}
\bibfield{author}{\bibinfo{person}{Haoran Chen}, \bibinfo{person}{Zuxuan Wu}, \bibinfo{person}{Xintong Han}, \bibinfo{person}{Menglin Jia}, {and} \bibinfo{person}{Yu-Gang Jiang}.} \bibinfo{year}{2025}\natexlab{}.
\newblock \showarticletitle{Promptfusion: Decoupling stability and plasticity for continual learning}. In \bibinfo{booktitle}{\emph{European Conference on Computer Vision}}. Springer, \bibinfo{pages}{196--212}.
\newblock


\bibitem[Chen et~al\mbox{.}(2020a)]%
        {chen2020canzsl_cv}
\bibfield{author}{\bibinfo{person}{Zhi Chen}, \bibinfo{person}{Jingjing Li}, \bibinfo{person}{Yadan Luo}, \bibinfo{person}{Zi Huang}, {and} \bibinfo{person}{Yang Yang}.} \bibinfo{year}{2020}\natexlab{a}.
\newblock \showarticletitle{Canzsl: Cycle-consistent adversarial networks for zero-shot learning from natural language}. In \bibinfo{booktitle}{\emph{Proceedings of the IEEE/CVF winter conference on applications of computer vision}}. \bibinfo{pages}{874--883}.
\newblock


\bibitem[Chen et~al\mbox{.}(2021)]%
        {chen2021semantics_cv}
\bibfield{author}{\bibinfo{person}{Zhi Chen}, \bibinfo{person}{Yadan Luo}, \bibinfo{person}{Ruihong Qiu}, \bibinfo{person}{Sen Wang}, \bibinfo{person}{Zi Huang}, \bibinfo{person}{Jingjing Li}, {and} \bibinfo{person}{Zheng Zhang}.} \bibinfo{year}{2021}\natexlab{}.
\newblock \showarticletitle{Semantics disentangling for generalized zero-shot learning}. In \bibinfo{booktitle}{\emph{Proceedings of the IEEE/CVF international conference on computer vision}}. \bibinfo{pages}{8712--8720}.
\newblock


\bibitem[Chen et~al\mbox{.}(2020b)]%
        {chen2020rethinking_cv}
\bibfield{author}{\bibinfo{person}{Zhi Chen}, \bibinfo{person}{Sen Wang}, \bibinfo{person}{Jingjing Li}, {and} \bibinfo{person}{Zi Huang}.} \bibinfo{year}{2020}\natexlab{b}.
\newblock \showarticletitle{Rethinking generative zero-shot learning: An ensemble learning perspective for recognising visual patches}. In \bibinfo{booktitle}{\emph{Proceedings of the 28th ACM international conference on multimedia}}. \bibinfo{pages}{3413--3421}.
\newblock


\bibitem[Chen et~al\mbox{.}(2024)]%
        {chen2024fastedit_cv}
\bibfield{author}{\bibinfo{person}{Zhi Chen}, \bibinfo{person}{Zecheng Zhao}, \bibinfo{person}{Yadan Luo}, {and} \bibinfo{person}{Zi Huang}.} \bibinfo{year}{2024}\natexlab{}.
\newblock \showarticletitle{FastEdit: Fast Text-Guided Single-Image Editing via Semantic-Aware Diffusion Fine-Tuning}.
\newblock \bibinfo{journal}{\emph{arXiv preprint arXiv:2408.03355}} (\bibinfo{year}{2024}).
\newblock


\bibitem[Covington et~al\mbox{.}(2016)]%
        {covington2016deep}
\bibfield{author}{\bibinfo{person}{Paul Covington}, \bibinfo{person}{Jay Adams}, {and} \bibinfo{person}{Emre Sargin}.} \bibinfo{year}{2016}\natexlab{}.
\newblock \showarticletitle{Deep neural networks for youtube recommendations}. In \bibinfo{booktitle}{\emph{Proceedings of the 10th ACM conference on recommender systems}}. \bibinfo{pages}{191--198}.
\newblock


\bibitem[Deng et~al\mbox{.}(2023)]%
        {deng2023promptswitch}
\bibfield{author}{\bibinfo{person}{Chaorui Deng}, \bibinfo{person}{Qi Chen}, \bibinfo{person}{Pengda Qin}, \bibinfo{person}{Da Chen}, {and} \bibinfo{person}{Qi Wu}.} \bibinfo{year}{2023}\natexlab{}.
\newblock \showarticletitle{Prompt switch: Efficient clip adaptation for text-video retrieval}. In \bibinfo{booktitle}{\emph{Proceedings of the IEEE/CVF International Conference on Computer Vision}}. \bibinfo{pages}{15648--15658}.
\newblock


\bibitem[Ding et~al\mbox{.}(2022)]%
        {ding2022VR_LwF}
\bibfield{author}{\bibinfo{person}{Yuxuan Ding}, \bibinfo{person}{Lingqiao Liu}, \bibinfo{person}{Chunna Tian}, \bibinfo{person}{Jingyuan Yang}, {and} \bibinfo{person}{Haoxuan Ding}.} \bibinfo{year}{2022}\natexlab{}.
\newblock \showarticletitle{Don't stop learning: Towards continual learning for the clip model}.
\newblock \bibinfo{journal}{\emph{arXiv preprint arXiv:2207.09248}} (\bibinfo{year}{2022}).
\newblock


\bibitem[Dubey et~al\mbox{.}(2024)]%
        {dubey2024llama}
\bibfield{author}{\bibinfo{person}{Abhimanyu Dubey}, \bibinfo{person}{Abhinav Jauhri}, \bibinfo{person}{Abhinav Pandey}, \bibinfo{person}{Abhishek Kadian}, \bibinfo{person}{Ahmad Al-Dahle}, \bibinfo{person}{Aiesha Letman}, \bibinfo{person}{Akhil Mathur}, \bibinfo{person}{Alan Schelten}, \bibinfo{person}{Amy Yang}, \bibinfo{person}{Angela Fan}, {et~al\mbox{.}}} \bibinfo{year}{2024}\natexlab{}.
\newblock \showarticletitle{The llama 3 herd of models}.
\newblock \bibinfo{journal}{\emph{arXiv preprint arXiv:2407.21783}} (\bibinfo{year}{2024}).
\newblock


\bibitem[Fang et~al\mbox{.}(2023)]%
        {fang2023uatvr}
\bibfield{author}{\bibinfo{person}{Bo Fang}, \bibinfo{person}{Wenhao Wu}, \bibinfo{person}{Chang Liu}, \bibinfo{person}{Yu Zhou}, \bibinfo{person}{Yuxin Song}, \bibinfo{person}{Weiping Wang}, \bibinfo{person}{Xiangbo Shu}, \bibinfo{person}{Xiangyang Ji}, {and} \bibinfo{person}{Jingdong Wang}.} \bibinfo{year}{2023}\natexlab{}.
\newblock \showarticletitle{Uatvr: Uncertainty-adaptive text-video retrieval}. In \bibinfo{booktitle}{\emph{Proceedings of the IEEE/CVF International Conference on Computer Vision}}. \bibinfo{pages}{13723--13733}.
\newblock


\bibitem[Fang et~al\mbox{.}(2021a)]%
        {fang2021clip2video_tvr}
\bibfield{author}{\bibinfo{person}{Han Fang}, \bibinfo{person}{Pengfei Xiong}, \bibinfo{person}{Luhui Xu}, {and} \bibinfo{person}{Yu Chen}.} \bibinfo{year}{2021}\natexlab{a}.
\newblock \showarticletitle{Clip2video: Mastering video-text retrieval via image clip}.
\newblock \bibinfo{journal}{\emph{arXiv preprint arXiv:2106.11097}} (\bibinfo{year}{2021}).
\newblock


\bibitem[Fang et~al\mbox{.}(2021b)]%
        {fang2021clip2video}
\bibfield{author}{\bibinfo{person}{Han Fang}, \bibinfo{person}{Pengfei Xiong}, \bibinfo{person}{Luhui Xu}, {and} \bibinfo{person}{Yu Chen}.} \bibinfo{year}{2021}\natexlab{b}.
\newblock \showarticletitle{Clip2video: Mastering video-text retrieval via image clip}.
\newblock \bibinfo{journal}{\emph{arXiv preprint arXiv:2106.11097}} (\bibinfo{year}{2021}).
\newblock


\bibitem[Fedus et~al\mbox{.}(2022)]%
        {fedus2022switchMoE}
\bibfield{author}{\bibinfo{person}{William Fedus}, \bibinfo{person}{Barret Zoph}, {and} \bibinfo{person}{Noam Shazeer}.} \bibinfo{year}{2022}\natexlab{}.
\newblock \showarticletitle{Switch transformers: Scaling to trillion parameter models with simple and efficient sparsity}.
\newblock \bibinfo{journal}{\emph{Journal of Machine Learning Research}}  \bibinfo{volume}{23} (\bibinfo{year}{2022}), \bibinfo{pages}{1--39}.
\newblock


\bibitem[French(1999)]%
        {french1999catastrophic}
\bibfield{author}{\bibinfo{person}{Robert~M French}.} \bibinfo{year}{1999}\natexlab{}.
\newblock \showarticletitle{Catastrophic forgetting in connectionist networks}.
\newblock \bibinfo{journal}{\emph{Trends in cognitive sciences}} (\bibinfo{year}{1999}), \bibinfo{pages}{128--135}.
\newblock


\bibitem[Garg et~al\mbox{.}(2023)]%
        {garg2023tic}
\bibfield{author}{\bibinfo{person}{Saurabh Garg}, \bibinfo{person}{Mehrdad Farajtabar}, \bibinfo{person}{Hadi Pouransari}, \bibinfo{person}{Raviteja Vemulapalli}, \bibinfo{person}{Sachin Mehta}, \bibinfo{person}{Oncel Tuzel}, \bibinfo{person}{Vaishaal Shankar}, {and} \bibinfo{person}{Fartash Faghri}.} \bibinfo{year}{2023}\natexlab{}.
\newblock \showarticletitle{Tic-clip: Continual training of clip models}.
\newblock \bibinfo{journal}{\emph{arXiv preprint arXiv:2310.16226}} (\bibinfo{year}{2023}).
\newblock


\bibitem[Gorti et~al\mbox{.}(2022)]%
        {gorti2022XPool}
\bibfield{author}{\bibinfo{person}{Satya~Krishna Gorti}, \bibinfo{person}{No{\"e}l Vouitsis}, \bibinfo{person}{Junwei Ma}, \bibinfo{person}{Keyvan Golestan}, \bibinfo{person}{Maksims Volkovs}, \bibinfo{person}{Animesh Garg}, {and} \bibinfo{person}{Guangwei Yu}.} \bibinfo{year}{2022}\natexlab{}.
\newblock \showarticletitle{X-pool: Cross-modal language-video attention for text-video retrieval}. In \bibinfo{booktitle}{\emph{Proceedings of the IEEE/CVF conference on computer vision and pattern recognition}}. \bibinfo{pages}{5006--5015}.
\newblock


\bibitem[Guan et~al\mbox{.}(2023)]%
        {guan2023pidro_tvr}
\bibfield{author}{\bibinfo{person}{Peiyan Guan}, \bibinfo{person}{Renjing Pei}, \bibinfo{person}{Bin Shao}, \bibinfo{person}{Jianzhuang Liu}, \bibinfo{person}{Weimian Li}, \bibinfo{person}{Jiaxi Gu}, \bibinfo{person}{Hang Xu}, \bibinfo{person}{Songcen Xu}, \bibinfo{person}{Youliang Yan}, {and} \bibinfo{person}{Edmund~Y Lam}.} \bibinfo{year}{2023}\natexlab{}.
\newblock \showarticletitle{Pidro: Parallel isomeric attention with dynamic routing for text-video retrieval}. In \bibinfo{booktitle}{\emph{Proceedings of the IEEE/CVF International Conference on Computer Vision}}. \bibinfo{pages}{11164--11173}.
\newblock


\bibitem[Guo et~al\mbox{.}(2019)]%
        {GuoYWCZH19_CL}
\bibfield{author}{\bibinfo{person}{Lei Guo}, \bibinfo{person}{Hongzhi Yin}, \bibinfo{person}{Qinyong Wang}, \bibinfo{person}{Tong Chen}, \bibinfo{person}{Alexander Zhou}, {and} \bibinfo{person}{Nguyen Quoc~Viet Hung}.} \bibinfo{year}{2019}\natexlab{}.
\newblock \showarticletitle{Streaming Session-based Recommendation}. In \bibinfo{booktitle}{\emph{{KDD}}}. \bibinfo{publisher}{{ACM}}, \bibinfo{pages}{1569--1577}.
\newblock


\bibitem[Gupta(2024)]%
        {gupta_youtube_searches}
\bibfield{author}{\bibinfo{person}{Lokesh Gupta}.} \bibinfo{year}{2024}\natexlab{}.
\newblock \bibinfo{title}{How many daily searches are made on {YouTube}?}
\newblock \bibinfo{howpublished}{LinkedIn Pulse}.
\newblock
\urldef\tempurl%
\url{https://www.linkedin.com/pulse/how-many-daily-searches-made-youtube-lokesh-gupta-m6orc/}
\showURL{%
\tempurl}


\bibitem[Hadsell et~al\mbox{.}(2020)]%
        {hadsell2020embracing}
\bibfield{author}{\bibinfo{person}{Raia Hadsell}, \bibinfo{person}{Dushyant Rao}, \bibinfo{person}{Andrei~A Rusu}, {and} \bibinfo{person}{Razvan Pascanu}.} \bibinfo{year}{2020}\natexlab{}.
\newblock \showarticletitle{Embracing change: Continual learning in deep neural networks}.
\newblock \bibinfo{journal}{\emph{Trends in cognitive sciences}} \bibinfo{volume}{24}, \bibinfo{number}{12} (\bibinfo{year}{2020}), \bibinfo{pages}{1028--1040}.
\newblock


\bibitem[He et~al\mbox{.}(2021)]%
        {he2021improving_tra_tvr}
\bibfield{author}{\bibinfo{person}{Feng He}, \bibinfo{person}{Qi Wang}, \bibinfo{person}{Zhifan Feng}, \bibinfo{person}{Wenbin Jiang}, \bibinfo{person}{Yajuan L{\"u}}, \bibinfo{person}{Yong Zhu}, {and} \bibinfo{person}{Xiao Tan}.} \bibinfo{year}{2021}\natexlab{}.
\newblock \showarticletitle{Improving video retrieval by adaptive margin}. In \bibinfo{booktitle}{\emph{Proceedings of the 44th international ACM SIGIR conference on research and development in information retrieval}}. \bibinfo{pages}{1359--1368}.
\newblock


\bibitem[Ho et~al\mbox{.}(2020)]%
        {ho2020denoising_cv}
\bibfield{author}{\bibinfo{person}{Jonathan Ho}, \bibinfo{person}{Ajay Jain}, {and} \bibinfo{person}{Pieter Abbeel}.} \bibinfo{year}{2020}\natexlab{}.
\newblock \showarticletitle{Denoising diffusion probabilistic models}.
\newblock \bibinfo{journal}{\emph{Advances in neural information processing systems}}  \bibinfo{volume}{33} (\bibinfo{year}{2020}), \bibinfo{pages}{6840--6851}.
\newblock


\bibitem[Hsu et~al\mbox{.}(2018)]%
        {hsu2018CIL}
\bibfield{author}{\bibinfo{person}{YenChang Hsu}, \bibinfo{person}{YenCheng Liu}, \bibinfo{person}{Anita Ramasamy}, {and} \bibinfo{person}{Zsolt Kira}.} \bibinfo{year}{2018}\natexlab{}.
\newblock \showarticletitle{Re-evaluating continual learning scenarios: A categorization and case for strong baselines. Continual Learning Workshop}. In \bibinfo{booktitle}{\emph{32nd Conference on Neural Information Processing Systems}}.
\newblock


\bibitem[Hu et~al\mbox{.}({[n.\,d.]})]%
        {hulora}
\bibfield{author}{\bibinfo{person}{Edward~J Hu}, \bibinfo{person}{Phillip Wallis}, \bibinfo{person}{Zeyuan Allen-Zhu}, \bibinfo{person}{Yuanzhi Li}, \bibinfo{person}{Shean Wang}, \bibinfo{person}{Lu Wang}, \bibinfo{person}{Weizhu Chen}, {et~al\mbox{.}}} \bibinfo{year}{[n.\,d.]}\natexlab{}.
\newblock \showarticletitle{LoRA: Low-Rank Adaptation of Large Language Models}. In \bibinfo{booktitle}{\emph{International Conference on Learning Representations}}.
\newblock


\bibitem[Jacobs et~al\mbox{.}(1991)]%
        {jacobs1991adaptiveMoE}
\bibfield{author}{\bibinfo{person}{Robert~A Jacobs}, \bibinfo{person}{Michael~I Jordan}, \bibinfo{person}{Steven~J Nowlan}, {and} \bibinfo{person}{Geoffrey~E Hinton}.} \bibinfo{year}{1991}\natexlab{}.
\newblock \showarticletitle{Adaptive mixtures of local experts}.
\newblock \bibinfo{journal}{\emph{Neural computation}}  \bibinfo{volume}{3} (\bibinfo{year}{1991}), \bibinfo{pages}{79--87}.
\newblock


\bibitem[Jin et~al\mbox{.}(2023)]%
        {jin2023diffusionret_tvr}
\bibfield{author}{\bibinfo{person}{Peng Jin}, \bibinfo{person}{Hao Li}, \bibinfo{person}{Zesen Cheng}, \bibinfo{person}{Kehan Li}, \bibinfo{person}{Xiangyang Ji}, \bibinfo{person}{Chang Liu}, \bibinfo{person}{Li Yuan}, {and} \bibinfo{person}{Jie Chen}.} \bibinfo{year}{2023}\natexlab{}.
\newblock \showarticletitle{Diffusionret: Generative text-video retrieval with diffusion model}. In \bibinfo{booktitle}{\emph{Proceedings of the IEEE/CVF international conference on computer vision}}. \bibinfo{pages}{2470--2481}.
\newblock


\bibitem[Johnson et~al\mbox{.}(2019)]%
        {johnson2019billion}
\bibfield{author}{\bibinfo{person}{Jeff Johnson}, \bibinfo{person}{Matthijs Douze}, {and} \bibinfo{person}{Herv{\'e} J{\'e}gou}.} \bibinfo{year}{2019}\natexlab{}.
\newblock \showarticletitle{Billion-scale similarity search with GPUs}.
\newblock \bibinfo{journal}{\emph{IEEE Transactions on Big Data}} \bibinfo{volume}{7}, \bibinfo{number}{3} (\bibinfo{year}{2019}), \bibinfo{pages}{535--547}.
\newblock


\bibitem[Jung et~al\mbox{.}(2023)]%
        {jung2023DAP}
\bibfield{author}{\bibinfo{person}{Dahuin Jung}, \bibinfo{person}{Dongyoon Han}, \bibinfo{person}{Jihwan Bang}, {and} \bibinfo{person}{Hwanjun Song}.} \bibinfo{year}{2023}\natexlab{}.
\newblock \showarticletitle{Generating instance-level prompts for rehearsal-free continual learning}. In \bibinfo{booktitle}{\emph{Proceedings of the IEEE/CVF International Conference on Computer Vision}}. \bibinfo{pages}{11847--11857}.
\newblock


\bibitem[Kirkpatrick et~al\mbox{.}(2017)]%
        {kirkpatrick2017EWC}
\bibfield{author}{\bibinfo{person}{James Kirkpatrick}, \bibinfo{person}{Razvan Pascanu}, \bibinfo{person}{Neil Rabinowitz}, \bibinfo{person}{Joel Veness}, \bibinfo{person}{Guillaume Desjardins}, \bibinfo{person}{Andrei~A Rusu}, \bibinfo{person}{Kieran Milan}, \bibinfo{person}{John Quan}, \bibinfo{person}{Tiago Ramalho}, \bibinfo{person}{Agnieszka Grabska-Barwinska}, {et~al\mbox{.}}} \bibinfo{year}{2017}\natexlab{}.
\newblock \showarticletitle{Overcoming catastrophic forgetting in neural networks}.
\newblock \bibinfo{journal}{\emph{Proceedings of the national academy of sciences}}  \bibinfo{volume}{114} (\bibinfo{year}{2017}), \bibinfo{pages}{3521--3526}.
\newblock


\bibitem[Li et~al\mbox{.}(2023)]%
        {li2023progressive_tvr}
\bibfield{author}{\bibinfo{person}{Pandeng Li}, \bibinfo{person}{Chen-Wei Xie}, \bibinfo{person}{Liming Zhao}, \bibinfo{person}{Hongtao Xie}, \bibinfo{person}{Jiannan Ge}, \bibinfo{person}{Yun Zheng}, \bibinfo{person}{Deli Zhao}, {and} \bibinfo{person}{Yongdong Zhang}.} \bibinfo{year}{2023}\natexlab{}.
\newblock \showarticletitle{Progressive spatio-temporal prototype matching for text-video retrieval}. In \bibinfo{booktitle}{\emph{Proceedings of the IEEE/CVF International Conference on Computer Vision}}. \bibinfo{pages}{4100--4110}.
\newblock


\bibitem[Li and Hoiem(2017)]%
        {li2017LwF}
\bibfield{author}{\bibinfo{person}{Zhizhong Li} {and} \bibinfo{person}{Derek Hoiem}.} \bibinfo{year}{2017}\natexlab{}.
\newblock \showarticletitle{Learning without forgetting}.
\newblock \bibinfo{journal}{\emph{IEEE transactions on pattern analysis and machine intelligence}}  \bibinfo{volume}{40} (\bibinfo{year}{2017}), \bibinfo{pages}{2935--2947}.
\newblock


\bibitem[Lin et~al\mbox{.}(2021)]%
        {lin2021clear}
\bibfield{author}{\bibinfo{person}{Zhiqiu Lin}, \bibinfo{person}{Jia Shi}, \bibinfo{person}{Deepak Pathak}, {and} \bibinfo{person}{Deva Ramanan}.} \bibinfo{year}{2021}\natexlab{}.
\newblock \showarticletitle{The clear benchmark: Continual learning on real-world imagery}. In \bibinfo{booktitle}{\emph{Thirty-fifth conference on neural information processing systems datasets and benchmarks track (round 2)}}.
\newblock


\bibitem[Liu et~al\mbox{.}(2022)]%
        {liu2022TS2-Net}
\bibfield{author}{\bibinfo{person}{Yuqi Liu}, \bibinfo{person}{Pengfei Xiong}, \bibinfo{person}{Luhui Xu}, \bibinfo{person}{Shengming Cao}, {and} \bibinfo{person}{Qin Jin}.} \bibinfo{year}{2022}\natexlab{}.
\newblock \showarticletitle{Ts2-net: Token shift and selection transformer for text-video retrieval}. In \bibinfo{booktitle}{\emph{European conference on computer vision}}. \bibinfo{pages}{319--335}.
\newblock


\bibitem[Lomonaco and Maltoni(2017)]%
        {lomonaco2017core50}
\bibfield{author}{\bibinfo{person}{Vincenzo Lomonaco} {and} \bibinfo{person}{Davide Maltoni}.} \bibinfo{year}{2017}\natexlab{}.
\newblock \showarticletitle{Core50: a new dataset and benchmark for continuous object recognition}. In \bibinfo{booktitle}{\emph{Conference on robot learning}}. PMLR, \bibinfo{pages}{17--26}.
\newblock


\bibitem[Lopez-Paz and Ranzato(2017)]%
        {lopez2017gradient_cl}
\bibfield{author}{\bibinfo{person}{David Lopez-Paz} {and} \bibinfo{person}{Marc'Aurelio Ranzato}.} \bibinfo{year}{2017}\natexlab{}.
\newblock \showarticletitle{Gradient episodic memory for continual learning}.
\newblock \bibinfo{journal}{\emph{Advances in neural information processing systems}}  \bibinfo{volume}{30} (\bibinfo{year}{2017}).
\newblock


\bibitem[Luo et~al\mbox{.}(2022)]%
        {luo2022clip4clip}
\bibfield{author}{\bibinfo{person}{Huaishao Luo}, \bibinfo{person}{Lei Ji}, \bibinfo{person}{Ming Zhong}, \bibinfo{person}{Yang Chen}, \bibinfo{person}{Wen Lei}, \bibinfo{person}{Nan Duan}, {and} \bibinfo{person}{Tianrui Li}.} \bibinfo{year}{2022}\natexlab{}.
\newblock \showarticletitle{Clip4clip: An empirical study of clip for end to end video clip retrieval and captioning}.
\newblock \bibinfo{journal}{\emph{Neurocomputing}}  \bibinfo{volume}{508} (\bibinfo{year}{2022}), \bibinfo{pages}{293--304}.
\newblock


\bibitem[Lyle et~al\mbox{.}(2023)]%
        {lyle2023understanding}
\bibfield{author}{\bibinfo{person}{Clare Lyle}, \bibinfo{person}{Zeyu Zheng}, \bibinfo{person}{Evgenii Nikishin}, \bibinfo{person}{Bernardo~Avila Pires}, \bibinfo{person}{Razvan Pascanu}, {and} \bibinfo{person}{Will Dabney}.} \bibinfo{year}{2023}\natexlab{}.
\newblock \showarticletitle{Understanding plasticity in neural networks}. In \bibinfo{booktitle}{\emph{International Conference on Machine Learning}}. PMLR, \bibinfo{pages}{23190--23211}.
\newblock


\bibitem[McClelland et~al\mbox{.}(1995)]%
        {mcclellcatastrophic}
\bibfield{author}{\bibinfo{person}{James~L McClelland}, \bibinfo{person}{Bruce~L McNaughton}, {and} \bibinfo{person}{Randall~C O'Reilly}.} \bibinfo{year}{1995}\natexlab{}.
\newblock \showarticletitle{Why there are complementary learning systems in the hippocampus and neocortex: insights from the successes and failures of connectionist models of learning and memory.}
\newblock \bibinfo{journal}{\emph{Psychological review}} (\bibinfo{year}{1995}), \bibinfo{pages}{419}.
\newblock


\bibitem[McCloskey and Cohen(1989)]%
        {mccloskey1989catastrophic}
\bibfield{author}{\bibinfo{person}{Michael McCloskey} {and} \bibinfo{person}{Neal~J Cohen}.} \bibinfo{year}{1989}\natexlab{}.
\newblock \showarticletitle{Catastrophic interference in connectionist networks: The sequential learning problem}.
\newblock In \bibinfo{booktitle}{\emph{Psychology of learning and motivation}}. \bibinfo{publisher}{Elsevier}, \bibinfo{pages}{109--165}.
\newblock


\bibitem[McDonnell et~al\mbox{.}(2024)]%
        {mcdonnell2024ranpac_ptm_cl}
\bibfield{author}{\bibinfo{person}{Mark~D McDonnell}, \bibinfo{person}{Dong Gong}, \bibinfo{person}{Amin Parvaneh}, \bibinfo{person}{Ehsan Abbasnejad}, {and} \bibinfo{person}{Anton van~den Hengel}.} \bibinfo{year}{2024}\natexlab{}.
\newblock \showarticletitle{Ranpac: Random projections and pre-trained models for continual learning}.
\newblock \bibinfo{journal}{\emph{Advances in Neural Information Processing Systems}}  \bibinfo{volume}{36} (\bibinfo{year}{2024}).
\newblock


\bibitem[Miech et~al\mbox{.}(2019)]%
        {miech2019howto100m}
\bibfield{author}{\bibinfo{person}{Antoine Miech}, \bibinfo{person}{Dimitri Zhukov}, \bibinfo{person}{Jean-Baptiste Alayrac}, \bibinfo{person}{Makarand Tapaswi}, \bibinfo{person}{Ivan Laptev}, {and} \bibinfo{person}{Josef Sivic}.} \bibinfo{year}{2019}\natexlab{}.
\newblock \showarticletitle{Howto100m: Learning a text-video embedding by watching hundred million narrated video clips}. In \bibinfo{booktitle}{\emph{Proceedings of the IEEE/CVF international conference on computer vision}}. \bibinfo{pages}{2630--2640}.
\newblock


\bibitem[Nicolas et~al\mbox{.}(2023)]%
        {nicolas2023mop_prompt}
\bibfield{author}{\bibinfo{person}{Julien Nicolas}, \bibinfo{person}{Florent Chiaroni}, \bibinfo{person}{Imtiaz Ziko}, \bibinfo{person}{Ola Ahmad}, \bibinfo{person}{Christian Desrosiers}, {and} \bibinfo{person}{Jose Dolz}.} \bibinfo{year}{2023}\natexlab{}.
\newblock \showarticletitle{MoP-CLIP: A mixture of prompt-tuned CLIP models for domain incremental learning}.
\newblock \bibinfo{journal}{\emph{arXiv preprint arXiv:2307.05707}} (\bibinfo{year}{2023}).
\newblock


\bibitem[Qiu et~al\mbox{.}(2020)]%
        {QiuYHC20_CL}
\bibfield{author}{\bibinfo{person}{Ruihong Qiu}, \bibinfo{person}{Hongzhi Yin}, \bibinfo{person}{Zi Huang}, {and} \bibinfo{person}{Tong Chen}.} \bibinfo{year}{2020}\natexlab{}.
\newblock \showarticletitle{{GAG:} Global Attributed Graph Neural Network for Streaming Session-based Recommendation}. In \bibinfo{booktitle}{\emph{{SIGIR}}}. \bibinfo{publisher}{{ACM}}, \bibinfo{pages}{669--678}.
\newblock


\bibitem[Radford et~al\mbox{.}(2021)]%
        {radford2021CLIP}
\bibfield{author}{\bibinfo{person}{Alec Radford}, \bibinfo{person}{Jong~Wook Kim}, \bibinfo{person}{Chris Hallacy}, \bibinfo{person}{Aditya Ramesh}, \bibinfo{person}{Gabriel Goh}, \bibinfo{person}{Sandhini Agarwal}, \bibinfo{person}{Girish Sastry}, \bibinfo{person}{Amanda Askell}, \bibinfo{person}{Pamela Mishkin}, \bibinfo{person}{Jack Clark}, {et~al\mbox{.}}} \bibinfo{year}{2021}\natexlab{}.
\newblock \showarticletitle{Learning transferable visual models from natural language supervision}. In \bibinfo{booktitle}{\emph{International conference on machine learning}}. \bibinfo{pages}{8748--8763}.
\newblock


\bibitem[Rebuffi et~al\mbox{.}(2017)]%
        {rebuffi2017icarl_cil}
\bibfield{author}{\bibinfo{person}{Sylvestre-Alvise Rebuffi}, \bibinfo{person}{Alexander Kolesnikov}, \bibinfo{person}{Georg Sperl}, {and} \bibinfo{person}{Christoph~H Lampert}.} \bibinfo{year}{2017}\natexlab{}.
\newblock \showarticletitle{icarl: Incremental classifier and representation learning}. In \bibinfo{booktitle}{\emph{Proceedings of the IEEE conference on Computer Vision and Pattern Recognition}}. \bibinfo{pages}{2001--2010}.
\newblock


\bibitem[Riemer et~al\mbox{.}(2018)]%
        {riemer2018learning_cl}
\bibfield{author}{\bibinfo{person}{Matthew Riemer}, \bibinfo{person}{Ignacio Cases}, \bibinfo{person}{Robert Ajemian}, \bibinfo{person}{Miao Liu}, \bibinfo{person}{Irina Rish}, \bibinfo{person}{Yuhai Tu}, {and} \bibinfo{person}{Gerald Tesauro}.} \bibinfo{year}{2018}\natexlab{}.
\newblock \showarticletitle{Learning to learn without forgetting by maximizing transfer and minimizing interference}.
\newblock \bibinfo{journal}{\emph{arXiv preprint arXiv:1810.11910}} (\bibinfo{year}{2018}).
\newblock


\bibitem[Roady et~al\mbox{.}(2020)]%
        {roady2020stream}
\bibfield{author}{\bibinfo{person}{Ryne Roady}, \bibinfo{person}{Tyler~L Hayes}, \bibinfo{person}{Hitesh Vaidya}, {and} \bibinfo{person}{Christopher Kanan}.} \bibinfo{year}{2020}\natexlab{}.
\newblock \showarticletitle{Stream-51: Streaming classification and novelty detection from videos}. In \bibinfo{booktitle}{\emph{Proceedings of the IEEE/CVF Conference on Computer Vision and Pattern Recognition Workshops}}. \bibinfo{pages}{228--229}.
\newblock


\bibitem[Shazeer et~al\mbox{.}(2017)]%
        {shazeer2017outrageouslMoE}
\bibfield{author}{\bibinfo{person}{Noam Shazeer}, \bibinfo{person}{Azalia Mirhoseini}, \bibinfo{person}{Krzysztof Maziarz}, \bibinfo{person}{Andy Davis}, \bibinfo{person}{Quoc Le}, \bibinfo{person}{Geoffrey Hinton}, {and} \bibinfo{person}{Jeff Dean}.} \bibinfo{year}{2017}\natexlab{}.
\newblock \showarticletitle{Outrageously large neural networks: The sparsely-gated mixture-of-experts layer}.
\newblock \bibinfo{journal}{\emph{arXiv preprint arXiv:1701.06538}} (\bibinfo{year}{2017}).
\newblock


\bibitem[Shin et~al\mbox{.}(2017)]%
        {shin2017continual_cl}
\bibfield{author}{\bibinfo{person}{Hanul Shin}, \bibinfo{person}{Jung~Kwon Lee}, \bibinfo{person}{Jaehong Kim}, {and} \bibinfo{person}{Jiwon Kim}.} \bibinfo{year}{2017}\natexlab{}.
\newblock \showarticletitle{Continual learning with deep generative replay}.
\newblock \bibinfo{journal}{\emph{Advances in neural information processing systems}}  \bibinfo{volume}{30} (\bibinfo{year}{2017}).
\newblock


\bibitem[Srinivasan et~al\mbox{.}(2022)]%
        {srinivasan2022climb}
\bibfield{author}{\bibinfo{person}{Tejas Srinivasan}, \bibinfo{person}{Ting-Yun Chang}, \bibinfo{person}{Leticia Pinto~Alva}, \bibinfo{person}{Georgios Chochlakis}, \bibinfo{person}{Mohammad Rostami}, {and} \bibinfo{person}{Jesse Thomason}.} \bibinfo{year}{2022}\natexlab{}.
\newblock \showarticletitle{Climb: A continual learning benchmark for vision-and-language tasks}.
\newblock \bibinfo{journal}{\emph{Advances in Neural Information Processing Systems}}  \bibinfo{volume}{35} (\bibinfo{year}{2022}), \bibinfo{pages}{29440--29453}.
\newblock


\bibitem[Tian et~al\mbox{.}(2024)]%
        {tian2024hydralora}
\bibfield{author}{\bibinfo{person}{Chunlin Tian}, \bibinfo{person}{Zhan Shi}, \bibinfo{person}{Zhijiang Guo}, \bibinfo{person}{Li Li}, {and} \bibinfo{person}{Chengzhong Xu}.} \bibinfo{year}{2024}\natexlab{}.
\newblock \showarticletitle{HydraLoRA: An Asymmetric LoRA Architecture for Efficient Fine-Tuning}. In \bibinfo{booktitle}{\emph{Advances in Neural Information Processing Systems (NeurIPS)}}.
\newblock


\bibitem[Tianqi~Tang(2024)]%
        {Tang2024vilcobench}
\bibfield{author}{\bibinfo{person}{Hao Xue Celso De Melo Flora~Salim Tianqi~Tang, Shohreh~Deldari}.} \bibinfo{year}{2024}\natexlab{}.
\newblock \showarticletitle{Vi{LC}o-Bench: {VI}deo Language {CO}ntinual learning Benchmark}. In \bibinfo{booktitle}{\emph{The Thirty-eight Conference on Neural Information Processing Systems Datasets and Benchmarks Track}}.
\newblock


\bibitem[Van~de Ven and Tolias(2019)]%
        {van2019CIL}
\bibfield{author}{\bibinfo{person}{Gido~M Van~de Ven} {and} \bibinfo{person}{Andreas~S Tolias}.} \bibinfo{year}{2019}\natexlab{}.
\newblock \showarticletitle{Three scenarios for continual learning}.
\newblock \bibinfo{journal}{\emph{arXiv preprint arXiv:1904.07734}} (\bibinfo{year}{2019}).
\newblock


\bibitem[Vaswani et~al\mbox{.}(2017)]%
        {vaswani2017attention_cv}
\bibfield{author}{\bibinfo{person}{Ashish Vaswani}, \bibinfo{person}{Noam Shazeer}, \bibinfo{person}{Niki Parmar}, \bibinfo{person}{Jakob Uszkoreit}, \bibinfo{person}{Llion Jones}, \bibinfo{person}{Aidan~N Gomez}, \bibinfo{person}{{\L}ukasz Kaiser}, {and} \bibinfo{person}{Illia Polosukhin}.} \bibinfo{year}{2017}\natexlab{}.
\newblock \showarticletitle{Attention is all you need}.
\newblock \bibinfo{journal}{\emph{Advances in neural information processing systems}}  \bibinfo{volume}{30} (\bibinfo{year}{2017}).
\newblock


\bibitem[Villa et~al\mbox{.}(2022)]%
        {villa2022vclimb}
\bibfield{author}{\bibinfo{person}{Andr{\'e}s Villa}, \bibinfo{person}{Kumail Alhamoud}, \bibinfo{person}{Victor Escorcia}, \bibinfo{person}{Fabian Caba}, \bibinfo{person}{Juan~Le{\'o}n Alc{\'a}zar}, {and} \bibinfo{person}{Bernard Ghanem}.} \bibinfo{year}{2022}\natexlab{}.
\newblock \showarticletitle{vclimb: A novel video class incremental learning benchmark}. In \bibinfo{booktitle}{\emph{Proceedings of the IEEE/CVF Conference on Computer Vision and Pattern Recognition}}. \bibinfo{pages}{19035--19044}.
\newblock


\bibitem[Villa et~al\mbox{.}(2023)]%
        {PIVOT_villa}
\bibfield{author}{\bibinfo{person}{Andr{\'{e}}s Villa}, \bibinfo{person}{Juan Le{\'{o}}n~Alc{\'{a}}zar}, \bibinfo{person}{Motasem Alfarra}, \bibinfo{person}{Kumail Alhamoud}, \bibinfo{person}{Julio Hurtado}, \bibinfo{person}{Fabian Caba~Heilbron}, \bibinfo{person}{Alvaro Soto}, {and} \bibinfo{person}{Bernard Ghanem}.} \bibinfo{year}{2023}\natexlab{}.
\newblock \showarticletitle{{PIVOT:} Prompting for Video Continual Learning}. In \bibinfo{booktitle}{\emph{Proceedings of the IEEE/CVF Conference on Computer Vision and Pattern Recognition (CVPR)}}.
\newblock


\bibitem[Wang et~al\mbox{.}(2024)]%
        {wang2024textmass}
\bibfield{author}{\bibinfo{person}{Jiamian Wang}, \bibinfo{person}{Guohao Sun}, \bibinfo{person}{Pichao Wang}, \bibinfo{person}{Dongfang Liu}, \bibinfo{person}{Sohail Dianat}, \bibinfo{person}{Majid Rabbani}, \bibinfo{person}{Raghuveer Rao}, {and} \bibinfo{person}{Zhiqiang Tao}.} \bibinfo{year}{2024}\natexlab{}.
\newblock \showarticletitle{Text Is MASS: Modeling as Stochastic Embedding for Text-Video Retrieval}. In \bibinfo{booktitle}{\emph{Proceedings of the IEEE/CVF Conference on Computer Vision and Pattern Recognition}}. \bibinfo{pages}{16551--16560}.
\newblock


\bibitem[Wang et~al\mbox{.}(2022a)]%
        {wang2022sprompt}
\bibfield{author}{\bibinfo{person}{Yabin Wang}, \bibinfo{person}{Zhiwu Huang}, {and} \bibinfo{person}{Xiaopeng Hong}.} \bibinfo{year}{2022}\natexlab{a}.
\newblock \showarticletitle{S-Prompts Learning with Pre-trained Transformers: An Occam's Razor for Domain Incremental Learning}. In \bibinfo{booktitle}{\emph{Conference on Neural Information Processing Systems (NeurIPS)}}.
\newblock


\bibitem[Wang et~al\mbox{.}(2023)]%
        {wang2023isolation_mm}
\bibfield{author}{\bibinfo{person}{Yabin Wang}, \bibinfo{person}{Zhiheng Ma}, \bibinfo{person}{Zhiwu Huang}, \bibinfo{person}{Yaowei Wang}, \bibinfo{person}{Zhou Su}, {and} \bibinfo{person}{Xiaopeng Hong}.} \bibinfo{year}{2023}\natexlab{}.
\newblock \showarticletitle{Isolation and impartial aggregation: A paradigm of incremental learning without interference}. In \bibinfo{booktitle}{\emph{Proceedings of the AAAI Conference on Artificial Intelligence}}, Vol.~\bibinfo{volume}{37}. \bibinfo{pages}{10209--10217}.
\newblock


\bibitem[Wang et~al\mbox{.}(2022b)]%
        {wang2022dualprompt}
\bibfield{author}{\bibinfo{person}{Zifeng Wang}, \bibinfo{person}{Zizhao Zhang}, \bibinfo{person}{Sayna Ebrahimi}, \bibinfo{person}{Ruoxi Sun}, \bibinfo{person}{Han Zhang}, \bibinfo{person}{Chen-Yu Lee}, \bibinfo{person}{Xiaoqi Ren}, \bibinfo{person}{Guolong Su}, \bibinfo{person}{Vincent Perot}, \bibinfo{person}{Jennifer Dy}, {et~al\mbox{.}}} \bibinfo{year}{2022}\natexlab{b}.
\newblock \showarticletitle{Dualprompt: Complementary prompting for rehearsal-free continual learning}. In \bibinfo{booktitle}{\emph{European Conference on Computer Vision}}. \bibinfo{pages}{631--648}.
\newblock


\bibitem[Wang et~al\mbox{.}(2022c)]%
        {wang2022L2P}
\bibfield{author}{\bibinfo{person}{Zifeng Wang}, \bibinfo{person}{Zizhao Zhang}, \bibinfo{person}{Chen-Yu Lee}, \bibinfo{person}{Han Zhang}, \bibinfo{person}{Ruoxi Sun}, \bibinfo{person}{Xiaoqi Ren}, \bibinfo{person}{Guolong Su}, \bibinfo{person}{Vincent Perot}, \bibinfo{person}{Jennifer Dy}, {and} \bibinfo{person}{Tomas Pfister}.} \bibinfo{year}{2022}\natexlab{c}.
\newblock \showarticletitle{Learning to prompt for continual learning}. In \bibinfo{booktitle}{\emph{Proceedings of the IEEE/CVF conference on computer vision and pattern recognition}}. \bibinfo{pages}{139--149}.
\newblock


\bibitem[Wortsman et~al\mbox{.}(2022)]%
        {wortsman2022wise}
\bibfield{author}{\bibinfo{person}{Mitchell Wortsman}, \bibinfo{person}{Gabriel Ilharco}, \bibinfo{person}{Jong~Wook Kim}, \bibinfo{person}{Mike Li}, \bibinfo{person}{Simon Kornblith}, \bibinfo{person}{Rebecca Roelofs}, \bibinfo{person}{Raphael~Gontijo Lopes}, \bibinfo{person}{Hannaneh Hajishirzi}, \bibinfo{person}{Ali Farhadi}, \bibinfo{person}{Hongseok Namkoong}, {et~al\mbox{.}}} \bibinfo{year}{2022}\natexlab{}.
\newblock \showarticletitle{Robust fine-tuning of zero-shot models}. In \bibinfo{booktitle}{\emph{Proceedings of the IEEE/CVF conference on computer vision and pattern recognition}}. \bibinfo{pages}{7959--7971}.
\newblock


\bibitem[Wu et~al\mbox{.}(2023a)]%
        {wu2023cap4video_tvr}
\bibfield{author}{\bibinfo{person}{Wenhao Wu}, \bibinfo{person}{Haipeng Luo}, \bibinfo{person}{Bo Fang}, \bibinfo{person}{Jingdong Wang}, {and} \bibinfo{person}{Wanli Ouyang}.} \bibinfo{year}{2023}\natexlab{a}.
\newblock \showarticletitle{Cap4video: What can auxiliary captions do for text-video retrieval?}. In \bibinfo{booktitle}{\emph{Proceedings of the IEEE/CVF Conference on Computer Vision and Pattern Recognition}}. \bibinfo{pages}{10704--10713}.
\newblock


\bibitem[Wu et~al\mbox{.}(2023b)]%
        {wu2023cap4video}
\bibfield{author}{\bibinfo{person}{Wenhao Wu}, \bibinfo{person}{Haipeng Luo}, \bibinfo{person}{Bo Fang}, \bibinfo{person}{Jingdong Wang}, {and} \bibinfo{person}{Wanli Ouyang}.} \bibinfo{year}{2023}\natexlab{b}.
\newblock \showarticletitle{Cap4video: What can auxiliary captions do for text-video retrieval?}. In \bibinfo{booktitle}{\emph{Proceedings of the IEEE/CVF Conference on Computer Vision and Pattern Recognition}}. \bibinfo{pages}{10704--10713}.
\newblock


\bibitem[Xu et~al\mbox{.}(2016)]%
        {xu2016msrvtt}
\bibfield{author}{\bibinfo{person}{Jun Xu}, \bibinfo{person}{Tao Mei}, \bibinfo{person}{Ting Yao}, {and} \bibinfo{person}{Yong Rui}.} \bibinfo{year}{2016}\natexlab{}.
\newblock \showarticletitle{Msr-vtt: A large video description dataset for bridging video and language}. In \bibinfo{booktitle}{\emph{Proceedings of the IEEE conference on computer vision and pattern recognition}}. \bibinfo{pages}{5288--5296}.
\newblock


\bibitem[Xue et~al\mbox{.}(2023)]%
        {xue2023clipvip}
\bibfield{author}{\bibinfo{person}{Hongwei Xue}, \bibinfo{person}{Yuchong Sun}, \bibinfo{person}{Bei Liu}, \bibinfo{person}{Jianlong Fu}, \bibinfo{person}{Ruihua Song}, \bibinfo{person}{Houqiang Li}, {and} \bibinfo{person}{Jiebo Luo}.} \bibinfo{year}{2023}\natexlab{}.
\newblock \showarticletitle{Clip-vip: Adapting pre-trained image-text model to video-language alignment}. In \bibinfo{booktitle}{\emph{The Eleventh International Conference on Learning Representations}}.
\newblock


\bibitem[Yadav et~al\mbox{.}(2023)]%
        {prompt_cl}
\bibfield{author}{\bibinfo{person}{Prateek Yadav}, \bibinfo{person}{Qing Sun}, \bibinfo{person}{Hantian Ding}, \bibinfo{person}{Xiaopeng Li}, \bibinfo{person}{Dejiao Zhang}, \bibinfo{person}{Ming Tan}, \bibinfo{person}{Xiaofei Ma}, \bibinfo{person}{Parminder Bhatia}, \bibinfo{person}{Ramesh Nallapati}, \bibinfo{person}{Murali~Krishna Ramanathan}, {et~al\mbox{.}}} \bibinfo{year}{2023}\natexlab{}.
\newblock \showarticletitle{Exploring continual learning for code generation models}.
\newblock \bibinfo{journal}{\emph{arXiv preprint arXiv:2307.02435}} (\bibinfo{year}{2023}).
\newblock


\bibitem[Yang et~al\mbox{.}(2024)]%
        {yang2024dgl_tvr}
\bibfield{author}{\bibinfo{person}{Xiangpeng Yang}, \bibinfo{person}{Linchao Zhu}, \bibinfo{person}{Xiaohan Wang}, {and} \bibinfo{person}{Yi Yang}.} \bibinfo{year}{2024}\natexlab{}.
\newblock \showarticletitle{DGL: Dynamic Global-Local Prompt Tuning for Text-Video Retrieval}. In \bibinfo{booktitle}{\emph{Proceedings of the AAAI Conference on Artificial Intelligence}}, Vol.~\bibinfo{volume}{38}. \bibinfo{pages}{6540--6548}.
\newblock


\bibitem[Yu et~al\mbox{.}(2024)]%
        {yu2024MoEAdapter}
\bibfield{author}{\bibinfo{person}{Jiazuo Yu}, \bibinfo{person}{Yunzhi Zhuge}, \bibinfo{person}{Lu Zhang}, \bibinfo{person}{Ping Hu}, \bibinfo{person}{Dong Wang}, \bibinfo{person}{Huchuan Lu}, {and} \bibinfo{person}{You He}.} \bibinfo{year}{2024}\natexlab{}.
\newblock \showarticletitle{Boosting continual learning of vision-language models via mixture-of-experts adapters}. In \bibinfo{booktitle}{\emph{Proceedings of the IEEE/CVF Conference on Computer Vision and Pattern Recognition}}. \bibinfo{pages}{23219--23230}.
\newblock


\bibitem[Yu et~al\mbox{.}(2018)]%
        {yu2018joint}
\bibfield{author}{\bibinfo{person}{Youngjae Yu}, \bibinfo{person}{Jongseok Kim}, {and} \bibinfo{person}{Gunhee Kim}.} \bibinfo{year}{2018}\natexlab{}.
\newblock \showarticletitle{A joint sequence fusion model for video question answering and retrieval}. In \bibinfo{booktitle}{\emph{Proceedings of the European conference on computer vision (ECCV)}}. \bibinfo{pages}{471--487}.
\newblock


\bibitem[Yuan et~al\mbox{.}(2022)]%
        {YuanYHCWC22_CL}
\bibfield{author}{\bibinfo{person}{Wei Yuan}, \bibinfo{person}{Hongzhi Yin}, \bibinfo{person}{Tieke He}, \bibinfo{person}{Tong Chen}, \bibinfo{person}{Qiufeng Wang}, {and} \bibinfo{person}{Lizhen Cui}.} \bibinfo{year}{2022}\natexlab{}.
\newblock \showarticletitle{Unified Question Generation with Continual Lifelong Learning}. In \bibinfo{booktitle}{\emph{{WWW}}}. \bibinfo{publisher}{{ACM}}, \bibinfo{pages}{871--881}.
\newblock


\bibitem[Zenke et~al\mbox{.}(2017)]%
        {zenke2017SI}
\bibfield{author}{\bibinfo{person}{Friedemann Zenke}, \bibinfo{person}{Ben Poole}, {and} \bibinfo{person}{Surya Ganguli}.} \bibinfo{year}{2017}\natexlab{}.
\newblock \showarticletitle{Continual learning through synaptic intelligence}. In \bibinfo{booktitle}{\emph{International conference on machine learning}}. \bibinfo{pages}{3987--3995}.
\newblock


\bibitem[Zhang et~al\mbox{.}(2023)]%
        {zhang2023slca_ptm_cl}
\bibfield{author}{\bibinfo{person}{Gengwei Zhang}, \bibinfo{person}{Liyuan Wang}, \bibinfo{person}{Guoliang Kang}, \bibinfo{person}{Ling Chen}, {and} \bibinfo{person}{Yunchao Wei}.} \bibinfo{year}{2023}\natexlab{}.
\newblock \showarticletitle{Slca: Slow learner with classifier alignment for continual learning on a pre-trained model}. In \bibinfo{booktitle}{\emph{Proceedings of the IEEE/CVF International Conference on Computer Vision}}. \bibinfo{pages}{19148--19158}.
\newblock


\bibitem[Zhao et~al\mbox{.}(2022)]%
        {zhao2022centerclip_tvr}
\bibfield{author}{\bibinfo{person}{Shuai Zhao}, \bibinfo{person}{Linchao Zhu}, \bibinfo{person}{Xiaohan Wang}, {and} \bibinfo{person}{Yi Yang}.} \bibinfo{year}{2022}\natexlab{}.
\newblock \showarticletitle{Centerclip: Token clustering for efficient text-video retrieval}. In \bibinfo{booktitle}{\emph{Proceedings of the 45th International ACM SIGIR Conference on Research and Development in Information Retrieval}}. \bibinfo{pages}{970--981}.
\newblock


\bibitem[Zheng et~al\mbox{.}(2023)]%
        {zheng2023ZSCL}
\bibfield{author}{\bibinfo{person}{Zangwei Zheng}, \bibinfo{person}{Mingyuan Ma}, \bibinfo{person}{Kai Wang}, \bibinfo{person}{Ziheng Qin}, \bibinfo{person}{Xiangyu Yue}, {and} \bibinfo{person}{Yang You}.} \bibinfo{year}{2023}\natexlab{}.
\newblock \showarticletitle{Preventing zero-shot transfer degradation in continual learning of vision-language models}. In \bibinfo{booktitle}{\emph{Proceedings of the IEEE/CVF International Conference on Computer Vision}}. \bibinfo{pages}{19125--19136}.
\newblock


\bibitem[Zhou et~al\mbox{.}(2024)]%
        {zhou2024revisiting_ptm_cl}
\bibfield{author}{\bibinfo{person}{Da-Wei Zhou}, \bibinfo{person}{Zi-Wen Cai}, \bibinfo{person}{Han-Jia Ye}, \bibinfo{person}{De-Chuan Zhan}, {and} \bibinfo{person}{Ziwei Liu}.} \bibinfo{year}{2024}\natexlab{}.
\newblock \showarticletitle{Revisiting class-incremental learning with pre-trained models: Generalizability and adaptivity are all you need}.
\newblock \bibinfo{journal}{\emph{International Journal of Computer Vision}} (\bibinfo{year}{2024}), \bibinfo{pages}{1--21}.
\newblock


\bibitem[Zhou et~al\mbox{.}(2023)]%
        {zhou2023learning_mm}
\bibfield{author}{\bibinfo{person}{Da-Wei Zhou}, \bibinfo{person}{Yuanhan Zhang}, \bibinfo{person}{Jingyi Ning}, \bibinfo{person}{Han-Jia Ye}, \bibinfo{person}{De-Chuan Zhan}, {and} \bibinfo{person}{Ziwei Liu}.} \bibinfo{year}{2023}\natexlab{}.
\newblock \showarticletitle{Learning without forgetting for vision-language models}.
\newblock \bibinfo{journal}{\emph{arXiv preprint arXiv:2305.19270}} (\bibinfo{year}{2023}).
\newblock


\end{thebibliography}

\end{document}